\documentclass[10pt]{article}

\newif\ifanonsubmission
\anonsubmissionfalse

\ifanonsubmission
  \usepackage{tmlr}
\else
  \usepackage[preprint]{tmlr}
\fi

\usepackage[utf8]{inputenc} %
\usepackage[T1]{fontenc}    %
\usepackage{hyperref}       %
\usepackage{url}            %
\usepackage{xurl}
\usepackage{booktabs}       %
\usepackage{amsfonts}       %
\usepackage{nicefrac}       %
\usepackage{microtype}      %
\usepackage{xcolor}         %

\newcommand{\srci}[2]{\begin{tabular}[c]{@{}c@{}}#1\\[-0.35ex]{\tiny[#2]}\end{tabular}}

\usepackage{graphicx}       %
\usepackage{enumitem}
\usepackage{amsmath}
\usepackage{algorithm2e}
\usepackage{multirow}
\usepackage{makecell}
\usepackage{placeins}
\usepackage{tikz}

\hypersetup{
pdftitle={TAVIS: A Benchmark for Egocentric Active Vision and Anticipatory Gaze in Imitation Learning},
pdfkeywords={active vision, imitation learning, benchmark, robot learning, vision-language-action (VLA) models, anticipatory gaze, humanoid robotics, egocentric vision, GALT, LeRobot, IsaacLab}
}

\newcommand{\repourl}{%
\ifanonsubmission
  https://anonymous.4open.science/r/tavis-F5D7%
\else
  https://github.com/spiglerg/tavis%
\fi
}

\title{TAVIS: A Benchmark for Egocentric Active Vision and Anticipatory Gaze in Imitation Learning}

\author{\name Giacomo Spigler \email g.spigler@tilburguniversity.edu \\
      \addr Department of Intelligent Systems\\
      Tilburg University, Netherlands}

\begin{document}

\maketitle

\begin{abstract}

Active vision -- where a policy controls its own gaze during manipulation -- has emerged as a key capability for imitation learning, with multiple independent systems demonstrating its benefits in the past year. Yet there is no shared benchmark to compare approaches or quantify what active vision contributes, on which task types, and under what conditions. We introduce TAVIS, evaluation infrastructure for active-vision imitation learning, with two complementary task suites -- TAVIS-Head (5 tasks, global search via pan/tilt necks) and TAVIS-Hands (3 tasks, local occlusion via wrist cameras) -- on two humanoid torso embodiments (GR1T2, Reachy2), built on IsaacLab. TAVIS provides three evaluation primitives: a paired headcam-vs-fixedcam protocol on identical demonstrations; GALT (Gaze-Action Lead Time), a novel metric grounded in cognitive science and HRI that quantifies anticipatory gaze in learned policies; and procedural ID/OOD splits. Baseline experiments with Diffusion Policy and $\pi_0$ reveal that (i) active vision generally helps, consistently across training runs, but benefits are task-conditional rather than uniform; (ii) multi-task policies degrade sharply under controlled distribution shifts on both suites; and (iii) imitation alone yields anticipatory gaze that lands on the task-relevant object, with lead times comparable to those of the human demonstrations, although head motion is less smooth than in the demonstrations, an effect of action chunking that success rate does not reveal.  Code and evaluation scripts are released at \url{\repourl}; demonstrations (LeRobot v3.0; ${\sim}2200$ episodes) and trained baselines at \url{https://huggingface.co/tavis-benchmark}.

\end{abstract}

\begin{figure}[h!]%
  \centering
  \includegraphics[width=\textwidth]{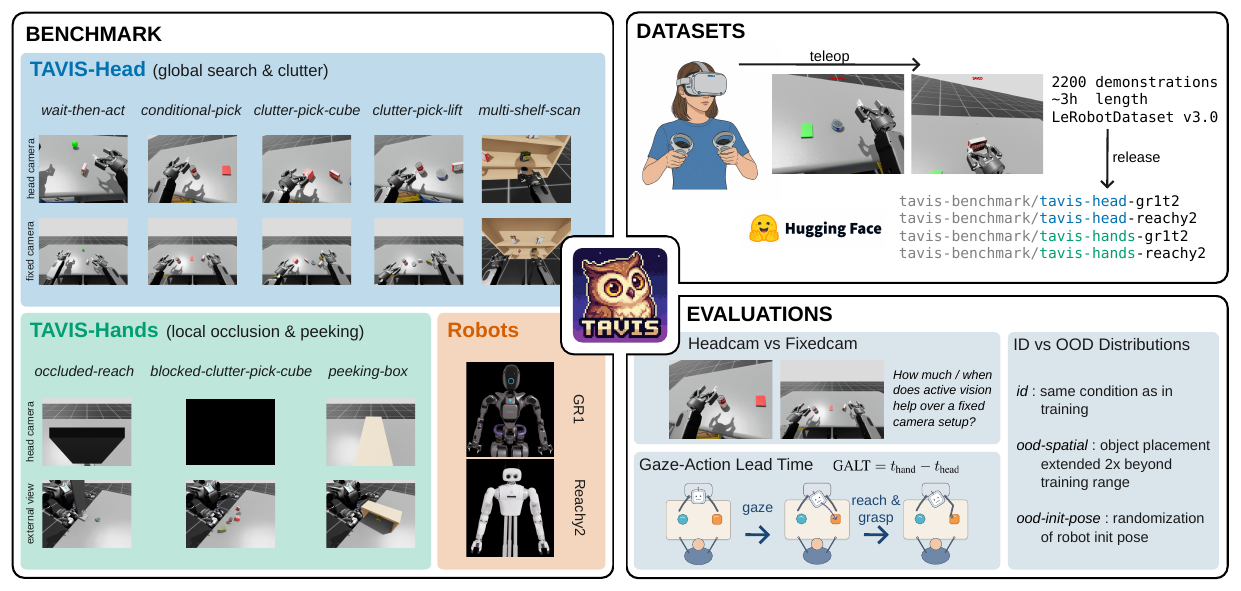}

  \caption{\textbf{The TAVIS Benchmark.} TAVIS comprises two task suites that isolate distinct roles of active vision in manipulation. \textit{TAVIS-Head} targets \emph{global} active vision -- head reorientation for search and to handle clutter -- while \textit{TAVIS-Hands} targets \emph{local} active vision via wrist cameras peering past occlusions. Demonstrations are collected via first-person Meta~Quest\,3 teleoperation with gaze control through head movements, and released on Hugging Face. The evaluation protocol spans three axes -- fixed vs head-mounted camera, in- vs out-of-distribution splits, and single- vs multi-task training -- along with the GALT (Gaze-Action Lead Time) metric for action legibility, enabling a thorough comparison of active-vision policies.} %
  \label{fig:teaser}
\end{figure}

\section{Introduction}

Imitation learning (IL) has progressed rapidly on visuomotor manipulation, yet most existing benchmarks and methods assume a fixed third-person or wide-angle camera. Over the past year, at least eight independent systems have shown that letting the policy control its own gaze can improve manipulation performance: ranging from high-DoF active necks~\citep{chuang2025active, xiong2025vision, yu2025egomi} to low-DoF stereo or eyeball heads~\citep{cheng2025open, kerr2025eye}, foveated and viewpoint-optimised vision~\citep{chuang2025look, liu2025avr}, and bimanual wrist-driven eyes~\citep{he2026towards}. These systems converge on the same finding: active gaze provides information that fixed cameras cannot, and policies that exploit it perform better.

Despite this convergence, no shared evaluation infrastructure exists for active-vision IL. Standard manipulation benchmarks --
LIBERO~\citep{liu2023libero}, RLBench~\citep{james2020rlbench}, CALVIN~\citep{mees2022calvin}, etc -- all assume fixed cameras and cannot isolate active vision as a controlled variable. Even EFM-10~\citep{he2026towards}, the closest existing artifact, is hardware-locked to a specific bimanual real-robot setup that other groups cannot easily reproduce. The eight systems above therefore cannot be meaningfully compared: each defines its own tasks, hardware, and metrics. Benchmarks have repeatedly catalyzed progress in machine learning by enabling fair comparison and identifying open problems; the absence of one for active-vision IL is a concrete bottleneck on the community's ability to assess what active vision contributes, on which task types, and under what conditions.

Indeed, active vision is a coupled perception-action problem: the policy must learn to control gaze in coordination with manipulation, and gaze itself serves multiple roles -- visual search, clutter disambiguation, temporal monitoring, and the communication of intent to human collaborators. This last role has deep precedent in cognitive science, where human gaze proactively leads the hand by hundreds of milliseconds at each contact landmark~\citep{johansson2001eye}, and in HRI, where this same temporal coupling makes a robot's behaviour \emph{legible} to bystanders~\citep{dragan2013legibility}. A policy that simply lifts the correct object scores identically to one that also uses anticipatory gaze to communicate intention. Evaluating active vision therefore requires new metrics designed to measure the temporal and communicative dimensions of gaze behaviour, not just task completion.

To close both gaps, we introduce the TAVIS benchmark (Figure~\ref{fig:teaser}) with two complementary task suites, \emph{TAVIS-Head} (global search and intent signalling through head gaze) and \emph{TAVIS-Hands} (local occlusion via wrist cameras), based on two humanoid torso embodiments (GR1T2, Reachy2).%

TAVIS is positioned as \emph{evaluation infrastructure} centered on three evaluations: a \emph{paired headcam-vs-fixedcam protocol} that isolates active vision as a controlled variable on identical demonstrations; \emph{GALT (Gaze-Action Lead Time)}, a novel metric that quantifies anticipatory gaze in successful episodes; and \emph{ID and OOD distribution splits} that distinguish in-distribution interpolation from extrapolation under controlled perturbations.%

\textbf{Overall, our novel contributions are:}
\begin{itemize}[leftmargin=*, itemsep=2pt, topsep=2pt]

    \item The TAVIS benchmark: 2 task suites (TAVIS-Head, TAVIS-Hands), 2 humanoid torsos (GR1T2, Reachy2), and a paired headcam/fixedcam evaluation protocol enabled by simultaneously-recorded demonstrations (LeRobot v3.0; ${\sim}2200$ episodes).

    \item Two evaluation primitives beyond task success: \emph{GALT}, a kinematic metric grounded in cognitive science and HRI for anticipatory gaze, validated against ground-truth gaze geometry and robust to its detector parameters; and \emph{ID/OOD distribution splits} extending the LIBERO-Pro~\citep{zhou2025liberoPRO} paradigm to active vision.

    \item Baseline analyses with Diffusion Policy~\citep{chi2023diffusionpolicy} and $\pi_0$~\citep{black2024pi0} showing (i) task-conditional active-vision benefits (+8 to +26pp headcam vs fixedcam at suite level on TAVIS-Head; 70–77\% success on TAVIS-Hands), (ii) sharp degradation under controlled OOD shifts on both suites, and (iii) anticipatory gaze acquired from imitation alone, directed at the target object with  lead times comparable to those of the human demonstrations, and a gaze-motion analysis that exposes less smooth head motion at action-chunk boundaries.

\end{itemize}

\section{Related Work}
\label{sec:related-work}

We organize related work along the three threads TAVIS draws on: imitation-learning benchmarks, active vision in manipulation, and gaze-coordination in cognitive science and HRI.

\paragraph{Imitation Learning Benchmarks.} Benchmarks have been instrumental to progress in machine learning by providing shared platforms for fair comparison; in robotics, where research ultimately targets hardware and sim-to-real gaps remain a real concern, simulation-based benchmark adoption has been more uneven. Nonetheless, some benchmarks have achieved significant impact in the field, including for example RLBench~\citep{james2020rlbench}, CALVIN~\citep{mees2022calvin}, RoboCasa~\citep{nasiriany2024robocasa}, RoboCerebra~\citep{hanrobocerebra}, and LIBERO~\citep{liu2023libero}. However, almost all of these benchmarks rely on fixed workspace cameras. The few exceptions that target active vision are EFM-10~\citep{he2026towards}, a real-robot benchmark for bimanual active perception, and AV-ALOHA~\citep{chuang2025active}, which provides teleoperation datasets and simulated environments. Outside of the visual domain, Tactile-MNIST~\citep{schneider2025tactile} provides a benchmark for active tactile perception. Beyond camera placement, most IL benchmarks are tied to a single robot platform, limiting cross-embodiment evaluation. TAVIS addresses both gaps in a single benchmark.

\paragraph{Active Vision in Robot Manipulation.} Active perception in robotics is a long-standing idea -- that an agent which controls its sensors can acquire information unavailable to a passive observer -- with foundational work by, e.g., \citet{bajcsy1988active}, \citet{aloimonos1988active}, and \citet{ballard1991animate}. Recently, the idea has been revisited from the perspective of imitation learning, whereby policies are trained to learn gaze directly from human teleoperation. The resulting systems vary mainly in how they organize the sensor's degrees of freedom. Low-DoF pan/tilt cameras -- mounted either on a humanoid neck, as in Open-Television~\citep{cheng2025open} and our TAVIS-Head setup, or on a mechanical eye gimbal, as in Eye Robot~\citep{kerr2025eye} -- provide the simplest configuration; higher-DoF active cameras can be realized either as a 6--7-DoF neck, as in AV-ALOHA~\citep{chuang2025active}, ViA~\citep{xiong2025vision}, and EgoMI~\citep{yu2025egomi}, or by using one arm as a movable eye, as in EFM-10~\citep{he2026towards} -- functionally similar despite the different embodiment. A complementary axis is image-space attention rather than camera motion: GIAVA~\citep{chuang2025look} extends AV-ALOHA with foveated processing driven by human eye-tracking, and AVR~\citep{liu2025avr} jointly optimizes viewpoint and focal length for precision tasks. Within this landscape, TAVIS-Head focuses on commodity pan/tilt necks, to capture most of the search and legibility benefits without the cost of a high-DoF neck, while TAVIS-Hands explores wrist-driven AV with \emph{both} arms used \emph{jointly} for perception and manipulation, in contrast to EFM-10's explicit one-arm-sees / one-arm-acts decomposition. %

\paragraph{Gaze Coordination and Legibility in Humans and Robots.} Two fields converge on the claim that gaze should precede action: cognitive-science studies of natural manipulation, and HRI work on legible motion. In natural object manipulation, gaze proactively marks upcoming contact landmarks rather than tracking the hand, leading the fingertips to each grasp or release site~\citep{johansson2001eye}. Reported lead times are $\sim$560\,ms in everyday tea-making~\citep{land1999roles} and $\sim$400\,ms in speed stacking~\citep{foerster2011saccadic}. This temporal coupling is sometimes formalized as the \emph{eye-hand arrival span}~\citep{kim2018temporal}, equivalent to the GALT metric we introduce in Section~\ref{sec:galt}.

In parallel, HRI work argues that \emph{how} a robot moves -- not only whether it succeeds -- shapes how readily humans interpret and accept its behaviour~\citep{dragan2013legibility}. Concretely, deliberately timed handovers let observers read the robot's gaze~\citep{admoni2014deliberate}, legible articulated pointing communicates intent~\citep{holladay2014legible}, and human-imitated head and gaze patterns improve naturalness on humanoid platforms~\citep{ding2024imitation}. Conversely, observers spontaneously generate anticipatory gaze toward robot goals~\citep{sciutti2013robots}, and recent learned models reproduce this gaze-primed reaching motion~\citep{hatano2025prime}.

Despite this convergent evidence, no IL benchmark has measured whether learned manipulation policies acquire anticipatory gaze; TAVIS introduces GALT (Section~\ref{sec:galt}) to close that gap, and validates it both on trained policies and on the human teleoperation reference.

\section{The TAVIS Benchmark}

TAVIS is a composable evaluation platform for egocentric active vision: two robot embodiments $\times$ two task suites $\times$ two camera modes (head-mounted vs fixed) $\times$ multiple distribution splits, all evaluable head-to-head. The two suites cover complementary regimes: \emph{TAVIS-Head} (global search via pan/tilt necks) and \emph{TAVIS-Hands} (local occlusion via wrist cameras). High-DoF necks are deliberately out of scope (Section~\ref{sec:related-work}).

We implemented TAVIS purely in simulation on top of IsaacLab~\citep{mittal2025isaaclab} and IsaacLab-Arena~\citep{isaaclabArena}, to take advantage of ray-traced rendering.

\subsection{Task Suites}
\label{sec:taskSuites}

Each task isolates one function of active vision, given in parentheses after its name below, so that performance differences between camera conditions can be read in terms of a single information need rather than compound planning or contact failures. Task design follows two conventions used throughout: unless otherwise specified, scenes draw from a fixed set of five distinct YCB objects, and language conditioning is task-dependent. Per-task scenes, prompts, randomization ranges, and success criteria are listed in Appendix~\ref{appendix:tasks_info}. Figure~\ref{fig:task-strips} shows one demonstration of each task. %

\noindent\textbf{TAVIS-Head.} The suite contains five tasks targeting roles where active head movement is expected to help:

\begin{itemize}[leftmargin=*, itemsep=1pt, topsep=1pt]
    \item \textit{conditional-pick} (conditional information gathering): Two objects are placed left and right; a colored card indicates the target (red = left, green = right). The robot looks at the card, then lifts the correct object. %

    \item \textit{wait-then-act} (temporal monitoring): The robot waits for a status light to turn from red to green after a randomized delay; only then it grasps and lifts the object. 

    \item \textit{clutter-pick-cube} (clutter disambiguation, target defined by appearance): A red cube and four distractor YCB objects are placed at randomized positions. The robot must locate the cube and lift it.

    \item \textit{clutter-pick-lift} (clutter disambiguation, target defined by language): A language prompt names a target among five objects. The robot must visually locate the object, grasp it, and lift it. For each object, 3 distinct prompts are used on different trials.

    \item \textit{multi-shelf-scan} (vertical workspace search): A three-shelf unit holds the target, which is named in a language prompt. The robot must scan the shelves vertically to locate it, then retrieve it. Like \textit{clutter-pick-lift}, each episode randomizes the instruction across 3 distinct prompts per object.
\end{itemize}

\noindent\textbf{TAVIS-Hands.} This suite targets local occlusion where head movement alone cannot reveal the target; the policy relies on wrist cameras for both perception and manipulation, using both arms jointly since the reachable hand is not known in advance. %

\begin{itemize}[leftmargin=*, itemsep=1pt, topsep=1pt]
    \item \textit{peeking-box} (looking into a container): A box with one side opening (left or right, randomized per episode) is placed on the table with a target object inside. The head camera cannot see the sides; the wrist cameras must determine which side is open. The robot reaches in with the corresponding hand and lifts the object.

    \item \textit{occluded-reach} (reaching around an occluder): A vertical screen sits on the table between the robot's head and the workspace, blocking the head's view of a single target object placed behind it. Wrist cameras provide the only useful view. The robot reaches around the screen and lifts the object.

    \item \textit{blocked-clutter-pick-cube} (wrist-only search): Identical to \textit{clutter-pick-cube}, except the robot's head camera is masked. The robot can only rely on the wrist cameras to locate and grasp the red cube. %
\end{itemize}

\begin{figure}[t]
  \centering
  \definecolor{galthead}{HTML}{1F77B4}\definecolor{galtgrasp}{HTML}{FF7F0E}
{\setlength{\tabcolsep}{0.5pt}\renewcommand{\arraystretch}{0.6}
\begin{tabular}{@{}r@{\hskip 3pt}cccccccc@{}}
\multicolumn{9}{@{}l}{\scriptsize\textbf{TAVIS-Head}} \\[1pt]
\begin{tabular}[c]{@{}r@{}}\scriptsize\textit{conditional-pick}\\\tiny head\end{tabular} & \tikz[baseline=(f.center)]{\node[inner sep=0pt] (f) {\includegraphics[width=0.105\textwidth]{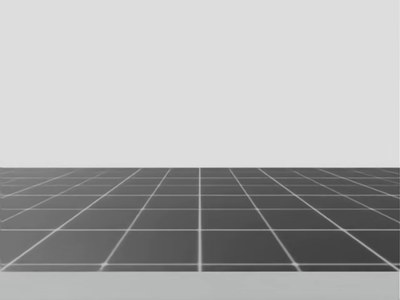}};\node[anchor=north west,inner sep=0.8pt,fill=white,fill opacity=0.75,text opacity=1,font=\fontsize{4.5}{5}\selectfont] at ([shift={(1.4pt,-1.4pt)}]f.north west) {0.0\,s};} & \tikz[baseline=(f.center)]{\node[inner sep=0pt] (f) {\includegraphics[width=0.105\textwidth]{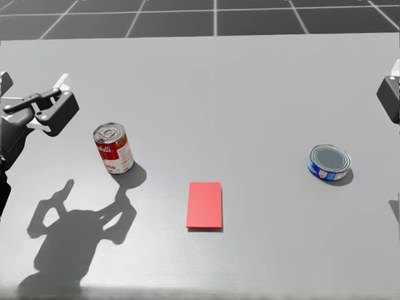}};\node[anchor=north west,inner sep=0.8pt,fill=white,fill opacity=0.75,text opacity=1,font=\fontsize{4.5}{5}\selectfont] at ([shift={(1.4pt,-1.4pt)}]f.north west) {1.0\,s};} & \tikz[baseline=(f.center)]{\node[inner sep=0pt] (f) {\includegraphics[width=0.105\textwidth]{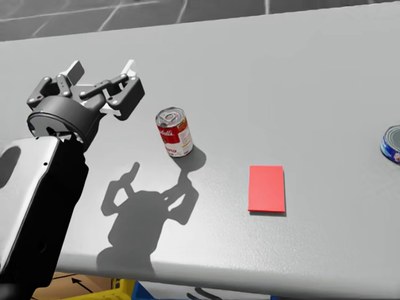}};\draw[galthead,line width=1.3pt] ([shift={(0.65pt,0.65pt)}]f.south west) rectangle ([shift={(-0.65pt,-0.65pt)}]f.north east);\node[anchor=north west,inner sep=0.8pt,fill=white,fill opacity=0.75,text opacity=1,font=\fontsize{4.5}{5}\selectfont] at ([shift={(1.4pt,-1.4pt)}]f.north west) {2.0\,s};} & \tikz[baseline=(f.center)]{\node[inner sep=0pt] (f) {\includegraphics[width=0.105\textwidth]{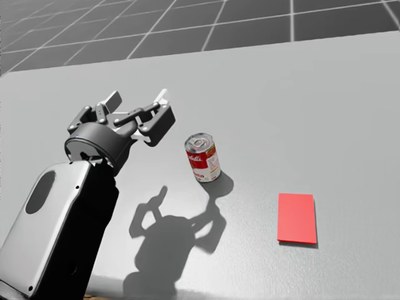}};\node[anchor=north west,inner sep=0.8pt,fill=white,fill opacity=0.75,text opacity=1,font=\fontsize{4.5}{5}\selectfont] at ([shift={(1.4pt,-1.4pt)}]f.north west) {2.6\,s};} & \tikz[baseline=(f.center)]{\node[inner sep=0pt] (f) {\includegraphics[width=0.105\textwidth]{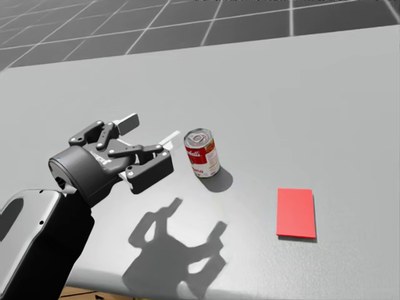}};\node[anchor=north west,inner sep=0.8pt,fill=white,fill opacity=0.75,text opacity=1,font=\fontsize{4.5}{5}\selectfont] at ([shift={(1.4pt,-1.4pt)}]f.north west) {3.3\,s};} & \tikz[baseline=(f.center)]{\node[inner sep=0pt] (f) {\includegraphics[width=0.105\textwidth]{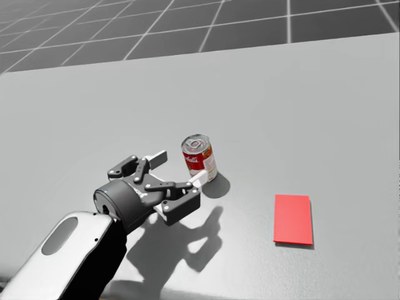}};\node[anchor=north west,inner sep=0.8pt,fill=white,fill opacity=0.75,text opacity=1,font=\fontsize{4.5}{5}\selectfont] at ([shift={(1.4pt,-1.4pt)}]f.north west) {3.9\,s};} & \tikz[baseline=(f.center)]{\node[inner sep=0pt] (f) {\includegraphics[width=0.105\textwidth]{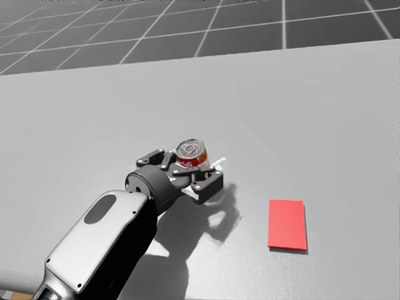}};\draw[galtgrasp,line width=1.3pt] ([shift={(0.65pt,0.65pt)}]f.south west) rectangle ([shift={(-0.65pt,-0.65pt)}]f.north east);\node[anchor=north west,inner sep=0.8pt,fill=white,fill opacity=0.75,text opacity=1,font=\fontsize{4.5}{5}\selectfont] at ([shift={(1.4pt,-1.4pt)}]f.north west) {4.5\,s};} & \tikz[baseline=(f.center)]{\node[inner sep=0pt] (f) {\includegraphics[width=0.105\textwidth]{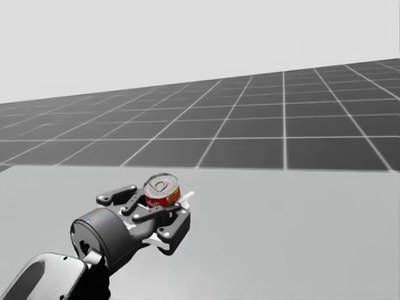}};\node[anchor=north west,inner sep=0.8pt,fill=white,fill opacity=0.75,text opacity=1,font=\fontsize{4.5}{5}\selectfont] at ([shift={(1.4pt,-1.4pt)}]f.north west) {6.2\,s};} \\[0.5pt]
\begin{tabular}[c]{@{}r@{}}\scriptsize\textit{wait-then-act}\\\tiny head\end{tabular} & \tikz[baseline=(f.center)]{\node[inner sep=0pt] (f) {\includegraphics[width=0.105\textwidth]{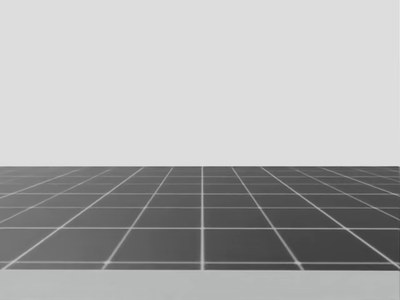}};\node[anchor=north west,inner sep=0.8pt,fill=white,fill opacity=0.75,text opacity=1,font=\fontsize{4.5}{5}\selectfont] at ([shift={(1.4pt,-1.4pt)}]f.north west) {0.0\,s};} & \tikz[baseline=(f.center)]{\node[inner sep=0pt] (f) {\includegraphics[width=0.105\textwidth]{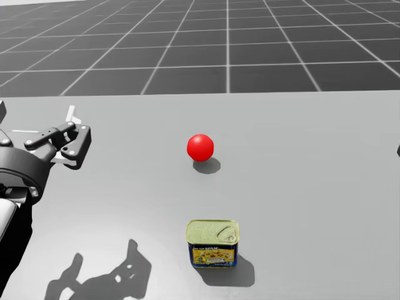}};\node[anchor=north west,inner sep=0.8pt,fill=white,fill opacity=0.75,text opacity=1,font=\fontsize{4.5}{5}\selectfont] at ([shift={(1.4pt,-1.4pt)}]f.north west) {2.1\,s};} & \tikz[baseline=(f.center)]{\node[inner sep=0pt] (f) {\includegraphics[width=0.105\textwidth]{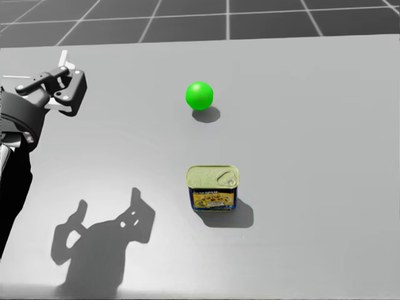}};\draw[galthead,line width=1.3pt] ([shift={(0.65pt,0.65pt)}]f.south west) rectangle ([shift={(-0.65pt,-0.65pt)}]f.north east);\node[anchor=north west,inner sep=0.8pt,fill=white,fill opacity=0.75,text opacity=1,font=\fontsize{4.5}{5}\selectfont] at ([shift={(1.4pt,-1.4pt)}]f.north west) {4.2\,s};} & \tikz[baseline=(f.center)]{\node[inner sep=0pt] (f) {\includegraphics[width=0.105\textwidth]{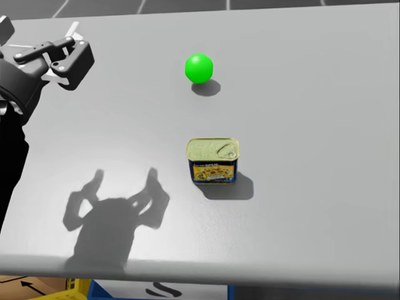}};\node[anchor=north west,inner sep=0.8pt,fill=white,fill opacity=0.75,text opacity=1,font=\fontsize{4.5}{5}\selectfont] at ([shift={(1.4pt,-1.4pt)}]f.north west) {4.8\,s};} & \tikz[baseline=(f.center)]{\node[inner sep=0pt] (f) {\includegraphics[width=0.105\textwidth]{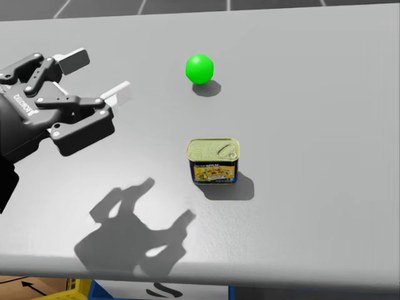}};\node[anchor=north west,inner sep=0.8pt,fill=white,fill opacity=0.75,text opacity=1,font=\fontsize{4.5}{5}\selectfont] at ([shift={(1.4pt,-1.4pt)}]f.north west) {5.5\,s};} & \tikz[baseline=(f.center)]{\node[inner sep=0pt] (f) {\includegraphics[width=0.105\textwidth]{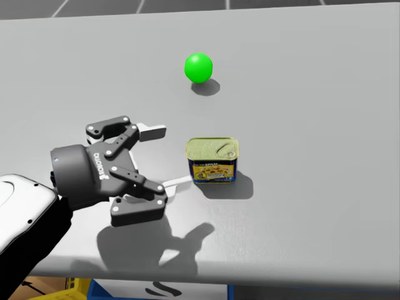}};\node[anchor=north west,inner sep=0.8pt,fill=white,fill opacity=0.75,text opacity=1,font=\fontsize{4.5}{5}\selectfont] at ([shift={(1.4pt,-1.4pt)}]f.north west) {6.1\,s};} & \tikz[baseline=(f.center)]{\node[inner sep=0pt] (f) {\includegraphics[width=0.105\textwidth]{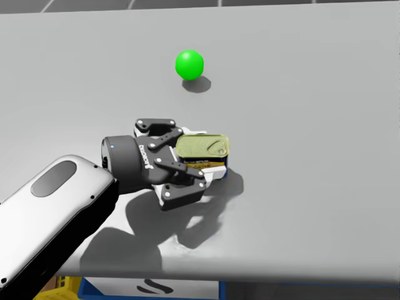}};\draw[galtgrasp,line width=1.3pt] ([shift={(0.65pt,0.65pt)}]f.south west) rectangle ([shift={(-0.65pt,-0.65pt)}]f.north east);\node[anchor=north west,inner sep=0.8pt,fill=white,fill opacity=0.75,text opacity=1,font=\fontsize{4.5}{5}\selectfont] at ([shift={(1.4pt,-1.4pt)}]f.north west) {6.7\,s};} & \tikz[baseline=(f.center)]{\node[inner sep=0pt] (f) {\includegraphics[width=0.105\textwidth]{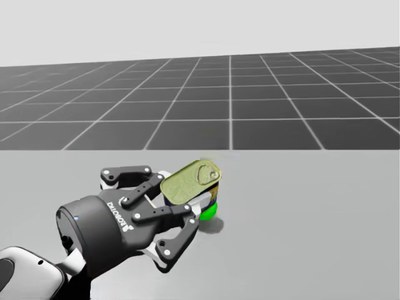}};\node[anchor=north west,inner sep=0.8pt,fill=white,fill opacity=0.75,text opacity=1,font=\fontsize{4.5}{5}\selectfont] at ([shift={(1.4pt,-1.4pt)}]f.north west) {8.7\,s};} \\[0.5pt]
\begin{tabular}[c]{@{}r@{}}\scriptsize\textit{clutter-pick-cube}\\\tiny head\end{tabular} & \tikz[baseline=(f.center)]{\node[inner sep=0pt] (f) {\includegraphics[width=0.105\textwidth]{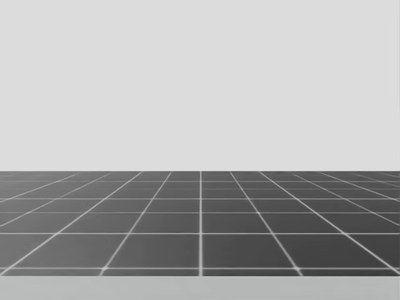}};\node[anchor=north west,inner sep=0.8pt,fill=white,fill opacity=0.75,text opacity=1,font=\fontsize{4.5}{5}\selectfont] at ([shift={(1.4pt,-1.4pt)}]f.north west) {0.0\,s};} & \tikz[baseline=(f.center)]{\node[inner sep=0pt] (f) {\includegraphics[width=0.105\textwidth]{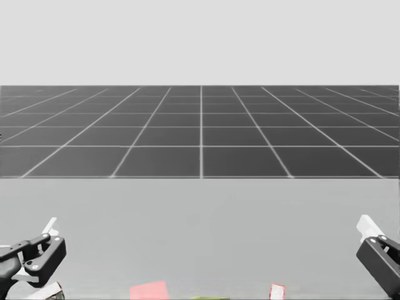}};\node[anchor=north west,inner sep=0.8pt,fill=white,fill opacity=0.75,text opacity=1,font=\fontsize{4.5}{5}\selectfont] at ([shift={(1.4pt,-1.4pt)}]f.north west) {0.6\,s};} & \tikz[baseline=(f.center)]{\node[inner sep=0pt] (f) {\includegraphics[width=0.105\textwidth]{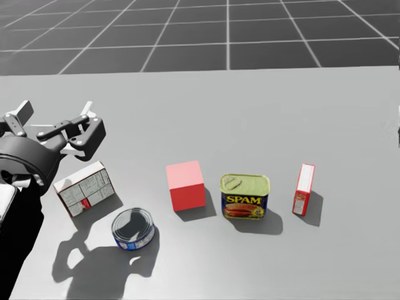}};\draw[galthead,line width=1.3pt] ([shift={(0.65pt,0.65pt)}]f.south west) rectangle ([shift={(-0.65pt,-0.65pt)}]f.north east);\node[anchor=north west,inner sep=0.8pt,fill=white,fill opacity=0.75,text opacity=1,font=\fontsize{4.5}{5}\selectfont] at ([shift={(1.4pt,-1.4pt)}]f.north west) {1.2\,s};} & \tikz[baseline=(f.center)]{\node[inner sep=0pt] (f) {\includegraphics[width=0.105\textwidth]{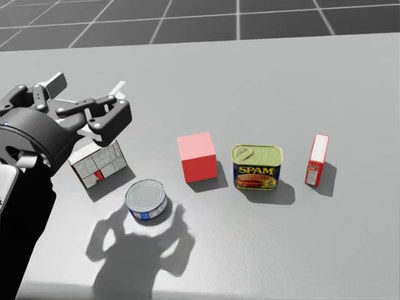}};\node[anchor=north west,inner sep=0.8pt,fill=white,fill opacity=0.75,text opacity=1,font=\fontsize{4.5}{5}\selectfont] at ([shift={(1.4pt,-1.4pt)}]f.north west) {1.9\,s};} & \tikz[baseline=(f.center)]{\node[inner sep=0pt] (f) {\includegraphics[width=0.105\textwidth]{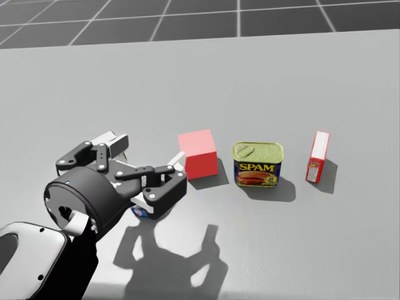}};\node[anchor=north west,inner sep=0.8pt,fill=white,fill opacity=0.75,text opacity=1,font=\fontsize{4.5}{5}\selectfont] at ([shift={(1.4pt,-1.4pt)}]f.north west) {2.5\,s};} & \tikz[baseline=(f.center)]{\node[inner sep=0pt] (f) {\includegraphics[width=0.105\textwidth]{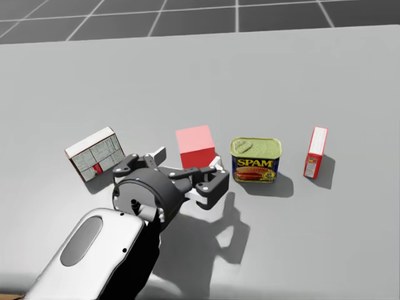}};\node[anchor=north west,inner sep=0.8pt,fill=white,fill opacity=0.75,text opacity=1,font=\fontsize{4.5}{5}\selectfont] at ([shift={(1.4pt,-1.4pt)}]f.north west) {3.2\,s};} & \tikz[baseline=(f.center)]{\node[inner sep=0pt] (f) {\includegraphics[width=0.105\textwidth]{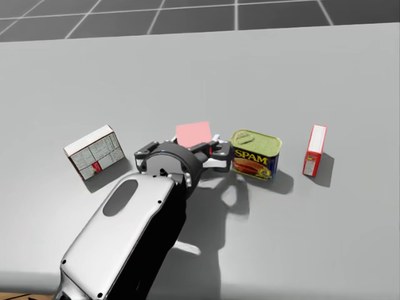}};\draw[galtgrasp,line width=1.3pt] ([shift={(0.65pt,0.65pt)}]f.south west) rectangle ([shift={(-0.65pt,-0.65pt)}]f.north east);\node[anchor=north west,inner sep=0.8pt,fill=white,fill opacity=0.75,text opacity=1,font=\fontsize{4.5}{5}\selectfont] at ([shift={(1.4pt,-1.4pt)}]f.north west) {3.9\,s};} & \tikz[baseline=(f.center)]{\node[inner sep=0pt] (f) {\includegraphics[width=0.105\textwidth]{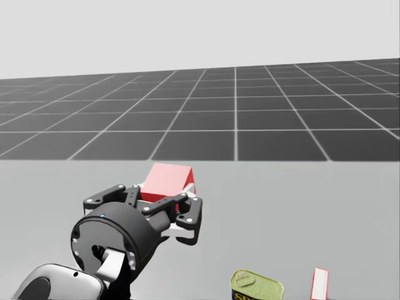}};\node[anchor=north west,inner sep=0.8pt,fill=white,fill opacity=0.75,text opacity=1,font=\fontsize{4.5}{5}\selectfont] at ([shift={(1.4pt,-1.4pt)}]f.north west) {5.9\,s};} \\[0.5pt]
\begin{tabular}[c]{@{}r@{}}\scriptsize\textit{clutter-pick-lift}\\\tiny head\end{tabular} & \tikz[baseline=(f.center)]{\node[inner sep=0pt] (f) {\includegraphics[width=0.105\textwidth]{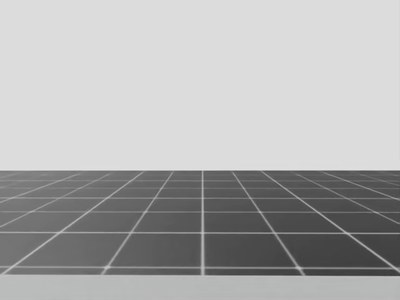}};\node[anchor=north west,inner sep=0.8pt,fill=white,fill opacity=0.75,text opacity=1,font=\fontsize{4.5}{5}\selectfont] at ([shift={(1.4pt,-1.4pt)}]f.north west) {0.0\,s};} & \tikz[baseline=(f.center)]{\node[inner sep=0pt] (f) {\includegraphics[width=0.105\textwidth]{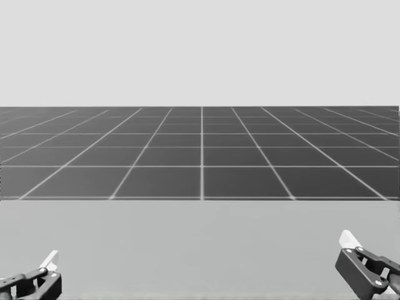}};\node[anchor=north west,inner sep=0.8pt,fill=white,fill opacity=0.75,text opacity=1,font=\fontsize{4.5}{5}\selectfont] at ([shift={(1.4pt,-1.4pt)}]f.north west) {0.5\,s};} & \tikz[baseline=(f.center)]{\node[inner sep=0pt] (f) {\includegraphics[width=0.105\textwidth]{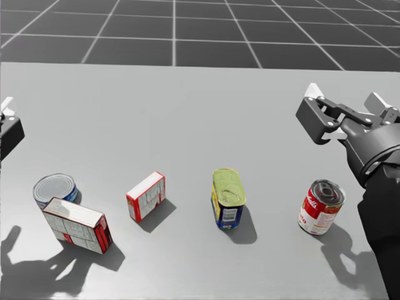}};\draw[galthead,line width=1.3pt] ([shift={(0.65pt,0.65pt)}]f.south west) rectangle ([shift={(-0.65pt,-0.65pt)}]f.north east);\node[anchor=north west,inner sep=0.8pt,fill=white,fill opacity=0.75,text opacity=1,font=\fontsize{4.5}{5}\selectfont] at ([shift={(1.4pt,-1.4pt)}]f.north west) {1.1\,s};} & \tikz[baseline=(f.center)]{\node[inner sep=0pt] (f) {\includegraphics[width=0.105\textwidth]{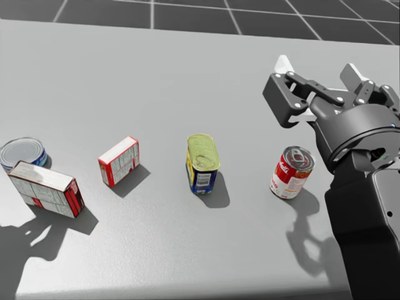}};\node[anchor=north west,inner sep=0.8pt,fill=white,fill opacity=0.75,text opacity=1,font=\fontsize{4.5}{5}\selectfont] at ([shift={(1.4pt,-1.4pt)}]f.north west) {1.7\,s};} & \tikz[baseline=(f.center)]{\node[inner sep=0pt] (f) {\includegraphics[width=0.105\textwidth]{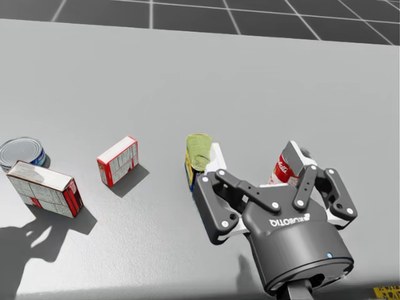}};\node[anchor=north west,inner sep=0.8pt,fill=white,fill opacity=0.75,text opacity=1,font=\fontsize{4.5}{5}\selectfont] at ([shift={(1.4pt,-1.4pt)}]f.north west) {2.3\,s};} & \tikz[baseline=(f.center)]{\node[inner sep=0pt] (f) {\includegraphics[width=0.105\textwidth]{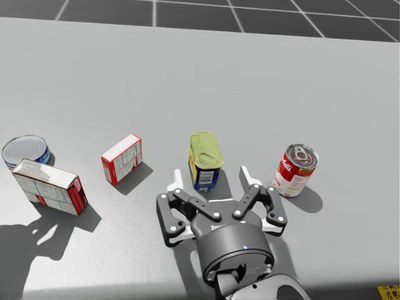}};\node[anchor=north west,inner sep=0.8pt,fill=white,fill opacity=0.75,text opacity=1,font=\fontsize{4.5}{5}\selectfont] at ([shift={(1.4pt,-1.4pt)}]f.north west) {3.0\,s};} & \tikz[baseline=(f.center)]{\node[inner sep=0pt] (f) {\includegraphics[width=0.105\textwidth]{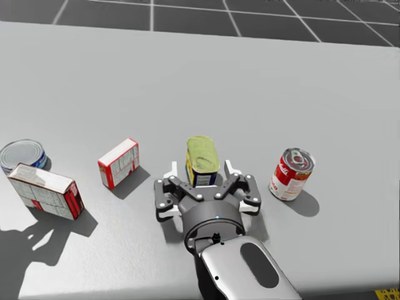}};\draw[galtgrasp,line width=1.3pt] ([shift={(0.65pt,0.65pt)}]f.south west) rectangle ([shift={(-0.65pt,-0.65pt)}]f.north east);\node[anchor=north west,inner sep=0.8pt,fill=white,fill opacity=0.75,text opacity=1,font=\fontsize{4.5}{5}\selectfont] at ([shift={(1.4pt,-1.4pt)}]f.north west) {3.6\,s};} & \tikz[baseline=(f.center)]{\node[inner sep=0pt] (f) {\includegraphics[width=0.105\textwidth]{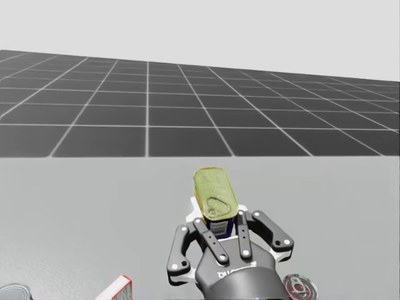}};\node[anchor=north west,inner sep=0.8pt,fill=white,fill opacity=0.75,text opacity=1,font=\fontsize{4.5}{5}\selectfont] at ([shift={(1.4pt,-1.4pt)}]f.north west) {5.3\,s};} \\[0.5pt]
\begin{tabular}[c]{@{}r@{}}\scriptsize\textit{multi-shelf-scan}\\\tiny head\end{tabular} & \tikz[baseline=(f.center)]{\node[inner sep=0pt] (f) {\includegraphics[width=0.105\textwidth]{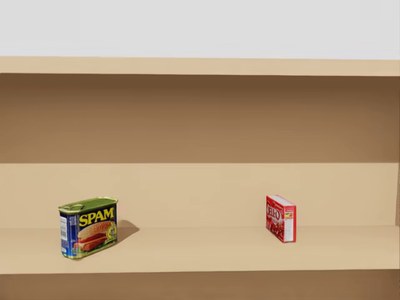}};\node[anchor=north west,inner sep=0.8pt,fill=white,fill opacity=0.75,text opacity=1,font=\fontsize{4.5}{5}\selectfont] at ([shift={(1.4pt,-1.4pt)}]f.north west) {0.0\,s};} & \tikz[baseline=(f.center)]{\node[inner sep=0pt] (f) {\includegraphics[width=0.105\textwidth]{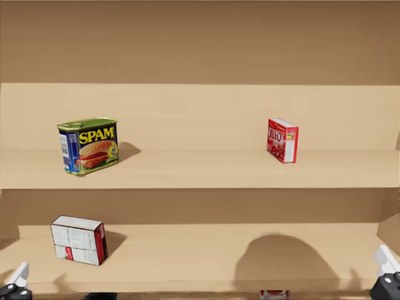}};\node[anchor=north west,inner sep=0.8pt,fill=white,fill opacity=0.75,text opacity=1,font=\fontsize{4.5}{5}\selectfont] at ([shift={(1.4pt,-1.4pt)}]f.north west) {0.5\,s};} & \tikz[baseline=(f.center)]{\node[inner sep=0pt] (f) {\includegraphics[width=0.105\textwidth]{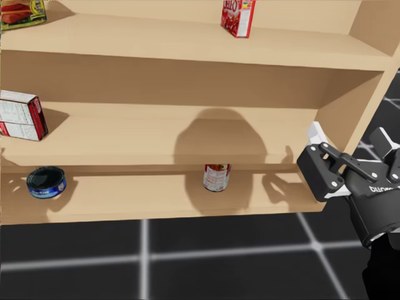}};\draw[galthead,line width=1.3pt] ([shift={(0.65pt,0.65pt)}]f.south west) rectangle ([shift={(-0.65pt,-0.65pt)}]f.north east);\node[anchor=north west,inner sep=0.8pt,fill=white,fill opacity=0.75,text opacity=1,font=\fontsize{4.5}{5}\selectfont] at ([shift={(1.4pt,-1.4pt)}]f.north west) {1.0\,s};} & \tikz[baseline=(f.center)]{\node[inner sep=0pt] (f) {\includegraphics[width=0.105\textwidth]{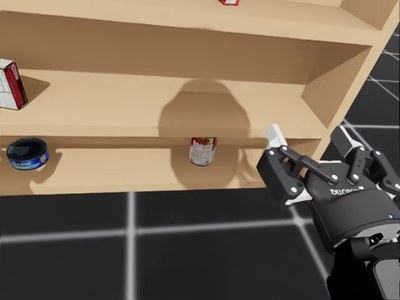}};\node[anchor=north west,inner sep=0.8pt,fill=white,fill opacity=0.75,text opacity=1,font=\fontsize{4.5}{5}\selectfont] at ([shift={(1.4pt,-1.4pt)}]f.north west) {1.7\,s};} & \tikz[baseline=(f.center)]{\node[inner sep=0pt] (f) {\includegraphics[width=0.105\textwidth]{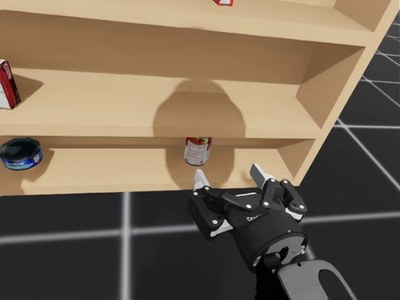}};\node[anchor=north west,inner sep=0.8pt,fill=white,fill opacity=0.75,text opacity=1,font=\fontsize{4.5}{5}\selectfont] at ([shift={(1.4pt,-1.4pt)}]f.north west) {2.4\,s};} & \tikz[baseline=(f.center)]{\node[inner sep=0pt] (f) {\includegraphics[width=0.105\textwidth]{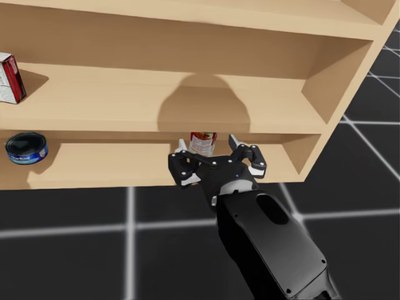}};\node[anchor=north west,inner sep=0.8pt,fill=white,fill opacity=0.75,text opacity=1,font=\fontsize{4.5}{5}\selectfont] at ([shift={(1.4pt,-1.4pt)}]f.north west) {3.1\,s};} & \tikz[baseline=(f.center)]{\node[inner sep=0pt] (f) {\includegraphics[width=0.105\textwidth]{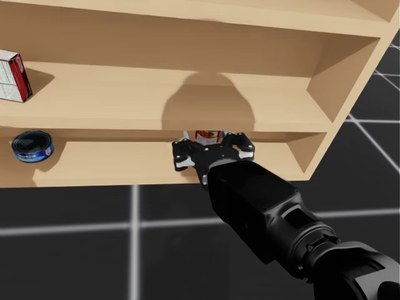}};\draw[galtgrasp,line width=1.3pt] ([shift={(0.65pt,0.65pt)}]f.south west) rectangle ([shift={(-0.65pt,-0.65pt)}]f.north east);\node[anchor=north west,inner sep=0.8pt,fill=white,fill opacity=0.75,text opacity=1,font=\fontsize{4.5}{5}\selectfont] at ([shift={(1.4pt,-1.4pt)}]f.north west) {3.7\,s};} & \tikz[baseline=(f.center)]{\node[inner sep=0pt] (f) {\includegraphics[width=0.105\textwidth]{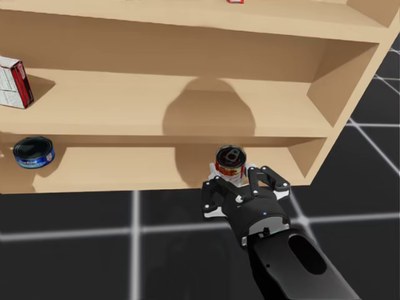}};\node[anchor=north west,inner sep=0.8pt,fill=white,fill opacity=0.75,text opacity=1,font=\fontsize{4.5}{5}\selectfont] at ([shift={(1.4pt,-1.4pt)}]f.north west) {4.7\,s};} \\[0.5pt]
\multicolumn{9}{@{}l}{\scriptsize\textbf{TAVIS-Hands}} \\[1pt]
\begin{tabular}[c]{@{}r@{}}\scriptsize\textit{peeking-box}\\\tiny wrist L\end{tabular} & \tikz[baseline=(f.center)]{\node[inner sep=0pt] (f) {\includegraphics[width=0.105\textwidth]{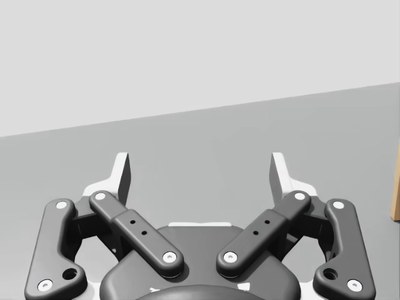}};\node[anchor=north west,inner sep=0.8pt,fill=white,fill opacity=0.75,text opacity=1,font=\fontsize{4.5}{5}\selectfont] at ([shift={(1.4pt,-1.4pt)}]f.north west) {0.0\,s};} & \tikz[baseline=(f.center)]{\node[inner sep=0pt] (f) {\includegraphics[width=0.105\textwidth]{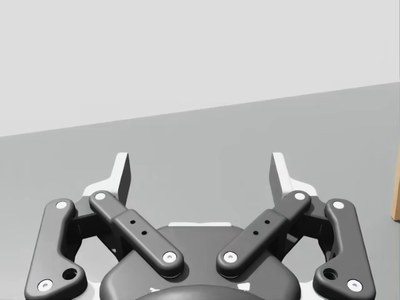}};\node[anchor=north west,inner sep=0.8pt,fill=white,fill opacity=0.75,text opacity=1,font=\fontsize{4.5}{5}\selectfont] at ([shift={(1.4pt,-1.4pt)}]f.north west) {0.4\,s};} & \tikz[baseline=(f.center)]{\node[inner sep=0pt] (f) {\includegraphics[width=0.105\textwidth]{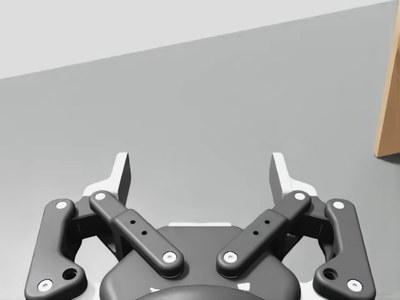}};\node[anchor=north west,inner sep=0.8pt,fill=white,fill opacity=0.75,text opacity=1,font=\fontsize{4.5}{5}\selectfont] at ([shift={(1.4pt,-1.4pt)}]f.north west) {0.7\,s};} & \tikz[baseline=(f.center)]{\node[inner sep=0pt] (f) {\includegraphics[width=0.105\textwidth]{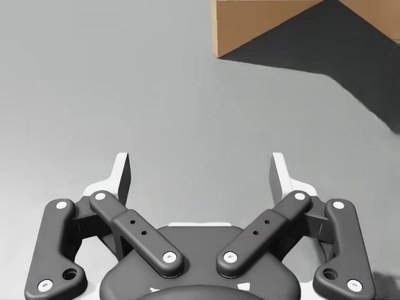}};\node[anchor=north west,inner sep=0.8pt,fill=white,fill opacity=0.75,text opacity=1,font=\fontsize{4.5}{5}\selectfont] at ([shift={(1.4pt,-1.4pt)}]f.north west) {1.1\,s};} & \tikz[baseline=(f.center)]{\node[inner sep=0pt] (f) {\includegraphics[width=0.105\textwidth]{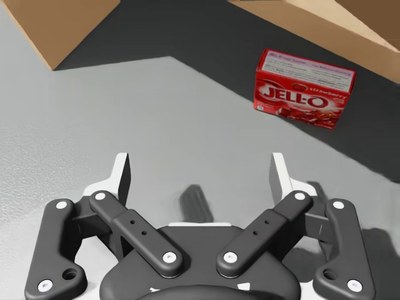}};\node[anchor=north west,inner sep=0.8pt,fill=white,fill opacity=0.75,text opacity=1,font=\fontsize{4.5}{5}\selectfont] at ([shift={(1.4pt,-1.4pt)}]f.north west) {1.4\,s};} & \tikz[baseline=(f.center)]{\node[inner sep=0pt] (f) {\includegraphics[width=0.105\textwidth]{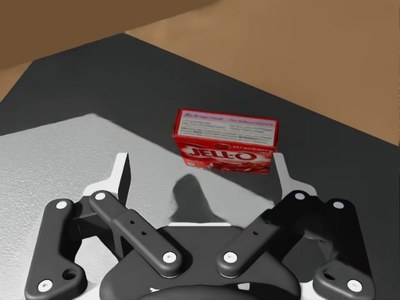}};\node[anchor=north west,inner sep=0.8pt,fill=white,fill opacity=0.75,text opacity=1,font=\fontsize{4.5}{5}\selectfont] at ([shift={(1.4pt,-1.4pt)}]f.north west) {1.8\,s};} & \tikz[baseline=(f.center)]{\node[inner sep=0pt] (f) {\includegraphics[width=0.105\textwidth]{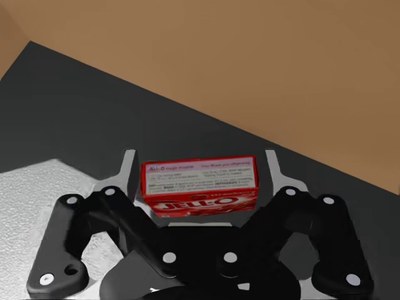}};\draw[galtgrasp,line width=1.3pt] ([shift={(0.65pt,0.65pt)}]f.south west) rectangle ([shift={(-0.65pt,-0.65pt)}]f.north east);\node[anchor=north west,inner sep=0.8pt,fill=white,fill opacity=0.75,text opacity=1,font=\fontsize{4.5}{5}\selectfont] at ([shift={(1.4pt,-1.4pt)}]f.north west) {2.1\,s};} & \tikz[baseline=(f.center)]{\node[inner sep=0pt] (f) {\includegraphics[width=0.105\textwidth]{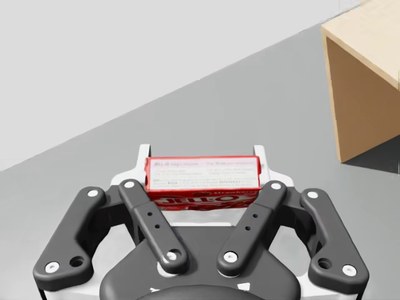}};\node[anchor=north west,inner sep=0.8pt,fill=white,fill opacity=0.75,text opacity=1,font=\fontsize{4.5}{5}\selectfont] at ([shift={(1.4pt,-1.4pt)}]f.north west) {3.3\,s};} \\[0.5pt]
\begin{tabular}[c]{@{}r@{}}\scriptsize\textit{occluded-reach}\\\tiny wrist L\end{tabular} & \tikz[baseline=(f.center)]{\node[inner sep=0pt] (f) {\includegraphics[width=0.105\textwidth]{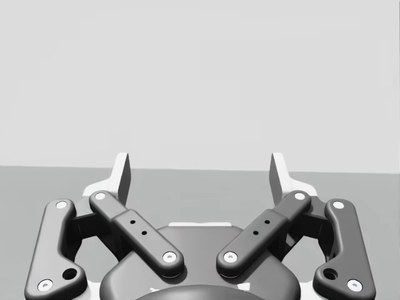}};\node[anchor=north west,inner sep=0.8pt,fill=white,fill opacity=0.75,text opacity=1,font=\fontsize{4.5}{5}\selectfont] at ([shift={(1.4pt,-1.4pt)}]f.north west) {0.0\,s};} & \tikz[baseline=(f.center)]{\node[inner sep=0pt] (f) {\includegraphics[width=0.105\textwidth]{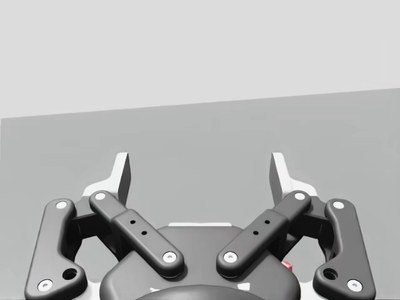}};\node[anchor=north west,inner sep=0.8pt,fill=white,fill opacity=0.75,text opacity=1,font=\fontsize{4.5}{5}\selectfont] at ([shift={(1.4pt,-1.4pt)}]f.north west) {0.5\,s};} & \tikz[baseline=(f.center)]{\node[inner sep=0pt] (f) {\includegraphics[width=0.105\textwidth]{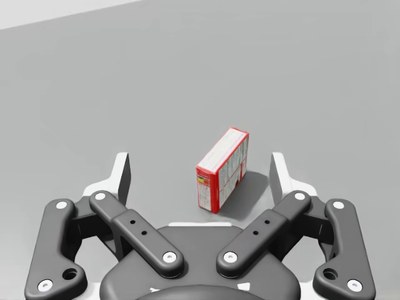}};\node[anchor=north west,inner sep=0.8pt,fill=white,fill opacity=0.75,text opacity=1,font=\fontsize{4.5}{5}\selectfont] at ([shift={(1.4pt,-1.4pt)}]f.north west) {1.0\,s};} & \tikz[baseline=(f.center)]{\node[inner sep=0pt] (f) {\includegraphics[width=0.105\textwidth]{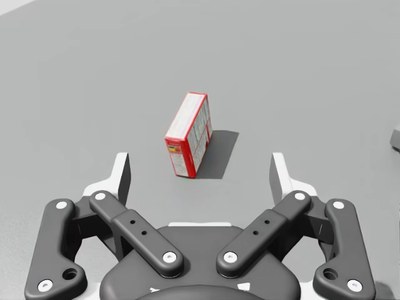}};\node[anchor=north west,inner sep=0.8pt,fill=white,fill opacity=0.75,text opacity=1,font=\fontsize{4.5}{5}\selectfont] at ([shift={(1.4pt,-1.4pt)}]f.north west) {1.5\,s};} & \tikz[baseline=(f.center)]{\node[inner sep=0pt] (f) {\includegraphics[width=0.105\textwidth]{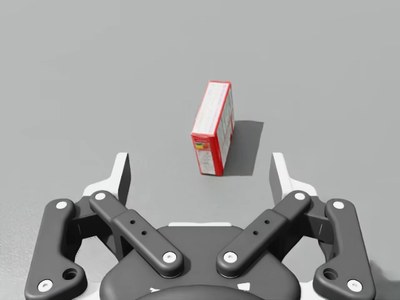}};\node[anchor=north west,inner sep=0.8pt,fill=white,fill opacity=0.75,text opacity=1,font=\fontsize{4.5}{5}\selectfont] at ([shift={(1.4pt,-1.4pt)}]f.north west) {2.0\,s};} & \tikz[baseline=(f.center)]{\node[inner sep=0pt] (f) {\includegraphics[width=0.105\textwidth]{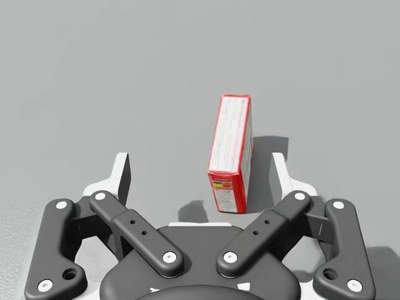}};\node[anchor=north west,inner sep=0.8pt,fill=white,fill opacity=0.75,text opacity=1,font=\fontsize{4.5}{5}\selectfont] at ([shift={(1.4pt,-1.4pt)}]f.north west) {2.5\,s};} & \tikz[baseline=(f.center)]{\node[inner sep=0pt] (f) {\includegraphics[width=0.105\textwidth]{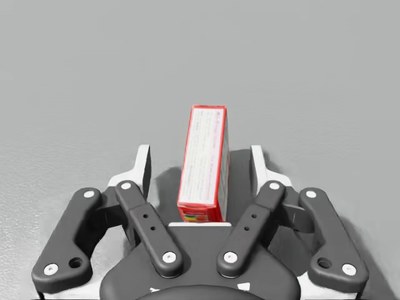}};\draw[galtgrasp,line width=1.3pt] ([shift={(0.65pt,0.65pt)}]f.south west) rectangle ([shift={(-0.65pt,-0.65pt)}]f.north east);\node[anchor=north west,inner sep=0.8pt,fill=white,fill opacity=0.75,text opacity=1,font=\fontsize{4.5}{5}\selectfont] at ([shift={(1.4pt,-1.4pt)}]f.north west) {3.0\,s};} & \tikz[baseline=(f.center)]{\node[inner sep=0pt] (f) {\includegraphics[width=0.105\textwidth]{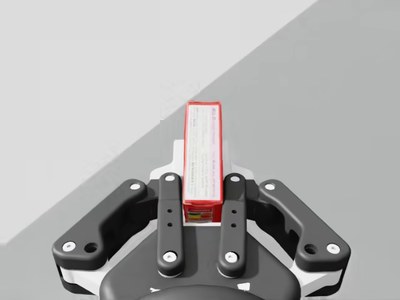}};\node[anchor=north west,inner sep=0.8pt,fill=white,fill opacity=0.75,text opacity=1,font=\fontsize{4.5}{5}\selectfont] at ([shift={(1.4pt,-1.4pt)}]f.north west) {4.5\,s};} \\[0.5pt]
\begin{tabular}[c]{@{}r@{}}\scriptsize\textit{blocked-clutter-}\\\scriptsize\textit{pick-cube}\\\tiny wrist L\end{tabular} & \tikz[baseline=(f.center)]{\node[inner sep=0pt] (f) {\includegraphics[width=0.105\textwidth]{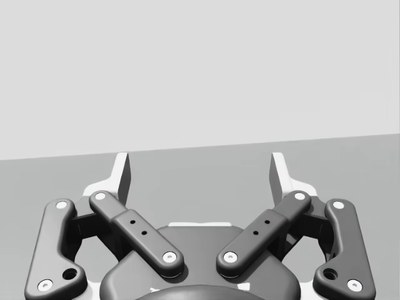}};\node[anchor=north west,inner sep=0.8pt,fill=white,fill opacity=0.75,text opacity=1,font=\fontsize{4.5}{5}\selectfont] at ([shift={(1.4pt,-1.4pt)}]f.north west) {0.0\,s};} & \tikz[baseline=(f.center)]{\node[inner sep=0pt] (f) {\includegraphics[width=0.105\textwidth]{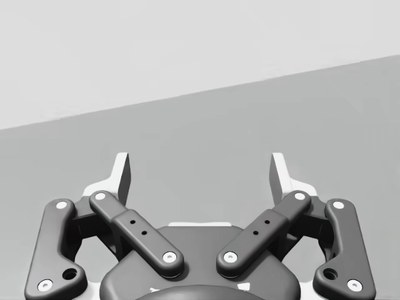}};\node[anchor=north west,inner sep=0.8pt,fill=white,fill opacity=0.75,text opacity=1,font=\fontsize{4.5}{5}\selectfont] at ([shift={(1.4pt,-1.4pt)}]f.north west) {0.6\,s};} & \tikz[baseline=(f.center)]{\node[inner sep=0pt] (f) {\includegraphics[width=0.105\textwidth]{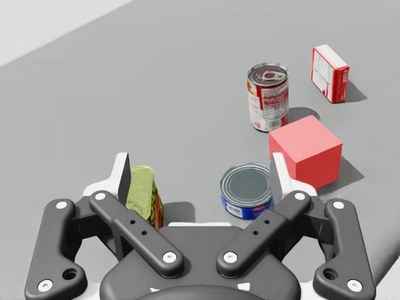}};\node[anchor=north west,inner sep=0.8pt,fill=white,fill opacity=0.75,text opacity=1,font=\fontsize{4.5}{5}\selectfont] at ([shift={(1.4pt,-1.4pt)}]f.north west) {1.2\,s};} & \tikz[baseline=(f.center)]{\node[inner sep=0pt] (f) {\includegraphics[width=0.105\textwidth]{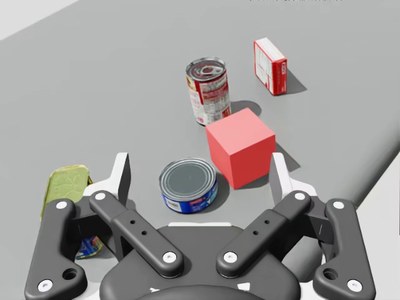}};\node[anchor=north west,inner sep=0.8pt,fill=white,fill opacity=0.75,text opacity=1,font=\fontsize{4.5}{5}\selectfont] at ([shift={(1.4pt,-1.4pt)}]f.north west) {1.8\,s};} & \tikz[baseline=(f.center)]{\node[inner sep=0pt] (f) {\includegraphics[width=0.105\textwidth]{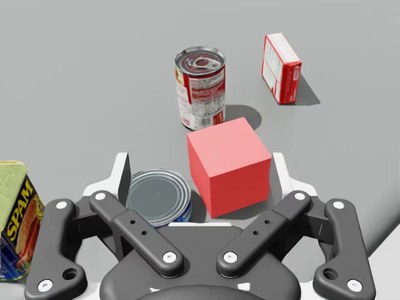}};\node[anchor=north west,inner sep=0.8pt,fill=white,fill opacity=0.75,text opacity=1,font=\fontsize{4.5}{5}\selectfont] at ([shift={(1.4pt,-1.4pt)}]f.north west) {2.4\,s};} & \tikz[baseline=(f.center)]{\node[inner sep=0pt] (f) {\includegraphics[width=0.105\textwidth]{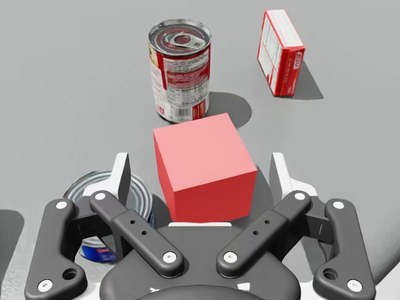}};\node[anchor=north west,inner sep=0.8pt,fill=white,fill opacity=0.75,text opacity=1,font=\fontsize{4.5}{5}\selectfont] at ([shift={(1.4pt,-1.4pt)}]f.north west) {3.0\,s};} & \tikz[baseline=(f.center)]{\node[inner sep=0pt] (f) {\includegraphics[width=0.105\textwidth]{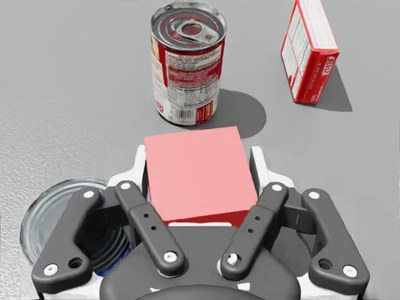}};\draw[galtgrasp,line width=1.3pt] ([shift={(0.65pt,0.65pt)}]f.south west) rectangle ([shift={(-0.65pt,-0.65pt)}]f.north east);\node[anchor=north west,inner sep=0.8pt,fill=white,fill opacity=0.75,text opacity=1,font=\fontsize{4.5}{5}\selectfont] at ([shift={(1.4pt,-1.4pt)}]f.north west) {3.6\,s};} & \tikz[baseline=(f.center)]{\node[inner sep=0pt] (f) {\includegraphics[width=0.105\textwidth]{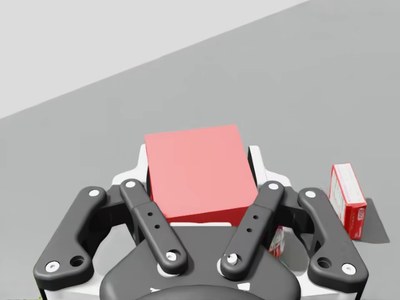}};\node[anchor=north west,inner sep=0.8pt,fill=white,fill opacity=0.75,text opacity=1,font=\fontsize{4.5}{5}\selectfont] at ([shift={(1.4pt,-1.4pt)}]f.north west) {4.8\,s};} \\[0.5pt]
\end{tabular}}

  \caption{\textbf{One demonstration per task (GR1T2).} Each row shows eight frames from one teleoperated demonstration, from the head camera for TAVIS-Head and from the wrist camera of the grasping arm for TAVIS-Hands. Columns are aligned to task events rather than spaced uniformly, so that the same column marks the same phase in every row. In TAVIS-Head, the blue frame is the head arrival at the final fixation and the orange frame is the grasp, so the interval between them is the GALT lead; in TAVIS-Hands, where head motion is not relevant, columns are aligned to the grasp only (orange).} %
  \label{fig:task-strips}
\end{figure}

\subsection{Robots}
\label{sec:robots}

TAVIS supports two fixed-base humanoid torsos, both with two 7-DoF arms and a 3-DoF neck: \textit{Reachy2} (Pollen Robotics) and \textit{GR1T2} (Fourier Intelligence, with Robotiq~2F-85 grippers replacing its native dexterous hands to match Reachy2).

Tasks operate on a unified 19-D canonical action space -- per-arm $\mathrm{SE}(3)$ end-effector target (xyz + unit quaternion), 3-DoF neck, per-hand scalar gripper -- defined in a hip-centred frame; full layout and the canonical-frame wrapper are in Appendix~\ref{appendix:robots}.

\subsection{Demonstration Collection and Datasets}
\label{sec:dataset}

Demonstrations are collected via a Quest~3 first-person-view teleoperation interface (Appendix~\ref{app:demonstrations_and_vr}). The operator views the head-camera feed (with optional side-by-side wrist-camera overlays), while controller pose drives the robot's bimanual end-effectors and headset orientation drives the neck. A central fixation marker keeps the operator's eye gaze stable, so head motion captures gaze.

A second \emph{fixed workspace camera} records the same episode in parallel, yielding paired head- and fixed-camera streams over identical trajectories -- the basis for the headcam-vs-fixedcam comparison in Section~\ref{sec:eval-paired}. Camera specs and control rates are in Appendix~\ref{appendix:robots}; teleop protocol and the LeRobot v3.0 release format with per-task and per-prompt fields are in Appendices~\ref{app:demonstrations_and_vr},~\ref{app:dataset_doc}. 

Datasets are released as four Hugging Face repositories\footnote{\url{https://huggingface.co/tavis-benchmark}}, one per (suite $\times$ robot) combination, totalling 2200 episodes. 

\section{Evaluation Protocol}

TAVIS provides three controlled evaluation axes: the \emph{camera mode} used by the policy (head-mounted vs fixed workspace camera, recorded simultaneously on identical demonstrations); the \emph{distribution split} (in-distribution and out-of-distribution variants); and the \emph{Gaze-Action Lead Time} (GALT), a kinematic metric that quantifies whether a policy's gaze anticipates its action. Throughout, we report success rate (SR) over 96 evaluation episodes, with Wilson $95\%$ confidence intervals in Appendix~\ref{app:suppl_results}; GALT is reported for successful TAVIS-Head episodes only.

\paragraph{Paired Headcam vs Fixedcam Comparison}
\label{sec:eval-paired}                                                                           
The central evaluation primitive of TAVIS is a \emph{paired} comparison between an agent-controlled head-mounted camera and a static workspace camera. Both image streams are recorded simultaneously from the same teleoperation episode (Section~\ref{sec:dataset}); any policy can therefore be trained and evaluated under either camera mode while every other variable -- demonstration, scene layout, robot trajectory, language prompt -- is held constant. We refer to the two resulting policy conditions as \emph{headcam} (head camera and both wrist cameras, with the neck controlled by the policy) and \emph{fixedcam} (fixed camera and both wrist cameras, with neck actions and neck state zeroed during both training and evaluation). In TAVIS-Head, both conditions can observe the whole workspace: the fixed camera sees all of it at once, with the wider field of view ($120^\circ$ vs about $70^\circ$ for the head camera), and the head camera by moving, which also lets it bring a target into a closer, more favourable view. TAVIS-Hands tasks omit the fixedcam condition by design: the head camera is structurally uninformative there, and a fixed camera adds no information the wrists do not already provide. %

A subtle confound is intrinsic to this paired setup: shared demonstrations cause fixedcam policies to inherit a brief `look-then-reach' pause from head-driven teleoperation. We treat this as ecologically valid (humans look before reaching) and quantify its magnitude (Section~\ref{sec:results}). %

\paragraph{ID and OOD Distribution Splits}
\label{sec:eval-splits}

TAVIS includes randomized OOD splits to test extrapolation, mitigating the memorization-not-generalization concern for fixed-configuration IL benchmarks raised by LIBERO-Pro \citep{zhou2025liberoPRO} and LIBERO-Plus \citep{fei2025liberoPLUS}. %

We define three cases. The \emph{in-distribution} (id) split samples within the same range used during demonstration collection, testing whether a policy interpolates over the variation it has already seen. \emph{ood-spatial} expands the distribution of initial positions of all objects to a larger region than in the training dataset. \emph{ood-init-pose} perturbs the robot reset pose (Gaussian noise $\sigma=10$cm on the Cartesian end-effector positions, $\sigma=10^\circ$ on the neck's yaw and pitch). Both perturbations change where things are (object placement and the robot's initial pose), and therefore the information-gathering problem itself; appearance-level shifts (lighting, textures, unseen objects), which mainly test visual generalization, are outside the scope of these splits. These ranges are intentionally aggressive: the same perturbation is applied uniformly across all checkpoints, so absolute success rates are biased downward but comparisons between methods under the same camera condition remain valid. %

\subsection{GALT: Gaze-Action Lead Time}
\label{sec:galt}

Cognitive science has long established that gaze precedes manual action: humans initiate eye fixations on a target $\sim$400-600\,ms before the corresponding hand movement \citep{johansson2001eye, land1999roles, foerster2011saccadic, kim2018temporal}. HRI work further shows that this temporal gap carries communicative value, allowing observers to read intent before action completion \citep{dragan2013legibility, admoni2014deliberate}. Despite the importance of this pattern, no robot-learning benchmark currently quantifies whether learned policies acquire it. We introduce GALT to fill this gap.

Note that TAVIS robots fixate via head and neck movements only, so absolute GALT values differ from the saccade literature and other AV platforms; we use the cog-sci range qualitatively to ground the anticipatory-gaze framing, not as a numerical target. %

\paragraph{Definition.}                                                                           
For a successful episode, let $t_{\text{head}}$ be the time at which the head-mounted camera reaches its final pre-grasp fixation, and $t_{\text{hand}}$ the time of grasp completion (gripper closure). We define

\begin{equation}
\text{GALT} = t_{\text{hand}} - t_{\text{head}},
\end{equation}

with $\text{GALT} > 0$ indicating anticipatory gaze. Using both events as \emph{arrivals} (rather than onsets) aligns with the ``eye-hand arrival span'' of \citet{kim2018temporal} and captures the legibility-relevant interval during which an observer can read intent and, in principle, intervene before contact.

GALT generalises to any action anchor (grasp, place, hand-off) and to multiple anchors per episode; here, each TAVIS task has a single grasp, so we report one value per successful episode. %

\paragraph{Detection.}

Both events are inferred from proprioception alone, making GALT portable to real-robot evaluation. The hand event $t_{\text{hand}}$ is the latest gripper-command transition (per-arm, with mutual-exclusion); $t_{\text{head}}$ is the arrival of the lowest-velocity neck fixation within a lookback window from $t_{\text{hand}}$. Simulation also allows checking the detected fixations against ground-truth geometry, which is not available on real robots. In our evaluation (Appendix~\ref{app:galt-algo}, Table~\ref{tab:galt-geometry}), the head camera's optical axis at the detected head arrival lies a median of 7.5$^\circ$ from the target object, compared with 52$^\circ$ at episode start, and the target is the object nearest to the gaze ray in 81\% of episodes, against 34\% expected by chance given the number of objects in each task. Full thresholds, exclusion codes, pseudocode, detection-rate validation, and a sensitivity analysis of the detector parameters appear in Appendix~\ref{app:galt-algo}. %

\section{Experiments and Analysis}
\label{sec:results}

We use TAVIS to investigate four questions about active-vision IL:

\begin{itemize}[leftmargin=*, itemsep=2pt, topsep=1pt]
    \item \textbf{Q1}: How much does active vision help, and on which task types? %

    \item \textbf{Q2}: To what extent does multi-task training help versus single-task training?  %

    \item \textbf{Q3}: What is the impact of distribution shift during evaluation? %

    \item \textbf{Q4}: Do policies acquire anticipatory gaze from imitation alone, and how does it compare to the human demonstrations?  %

\end{itemize}

\paragraph{Baselines and Training Details.}

We train two baselines via LeRobot~\citep{cadene2024lerobot} on each (suite, robot, camera-mode): $\pi_0$~\citep{black2024pi0} at single-task and suite-multi-task scopes (fine-tuned from \texttt{lerobot/pi0\_base}), and Diffusion Policy~\citep{chi2023diffusionpolicy} single-task on non-language-prompted tasks only. Each policy is evaluated for 96 episodes per condition; hyperparameters in Appendix~\ref{app:suppl_methods}. Multi-task $\pi_0$ checkpoints are released on Hugging Face\footnote{\url{https://huggingface.co/tavis-benchmark}}.

\begin{figure}[t]
    \centering
    \includegraphics[width=\textwidth]{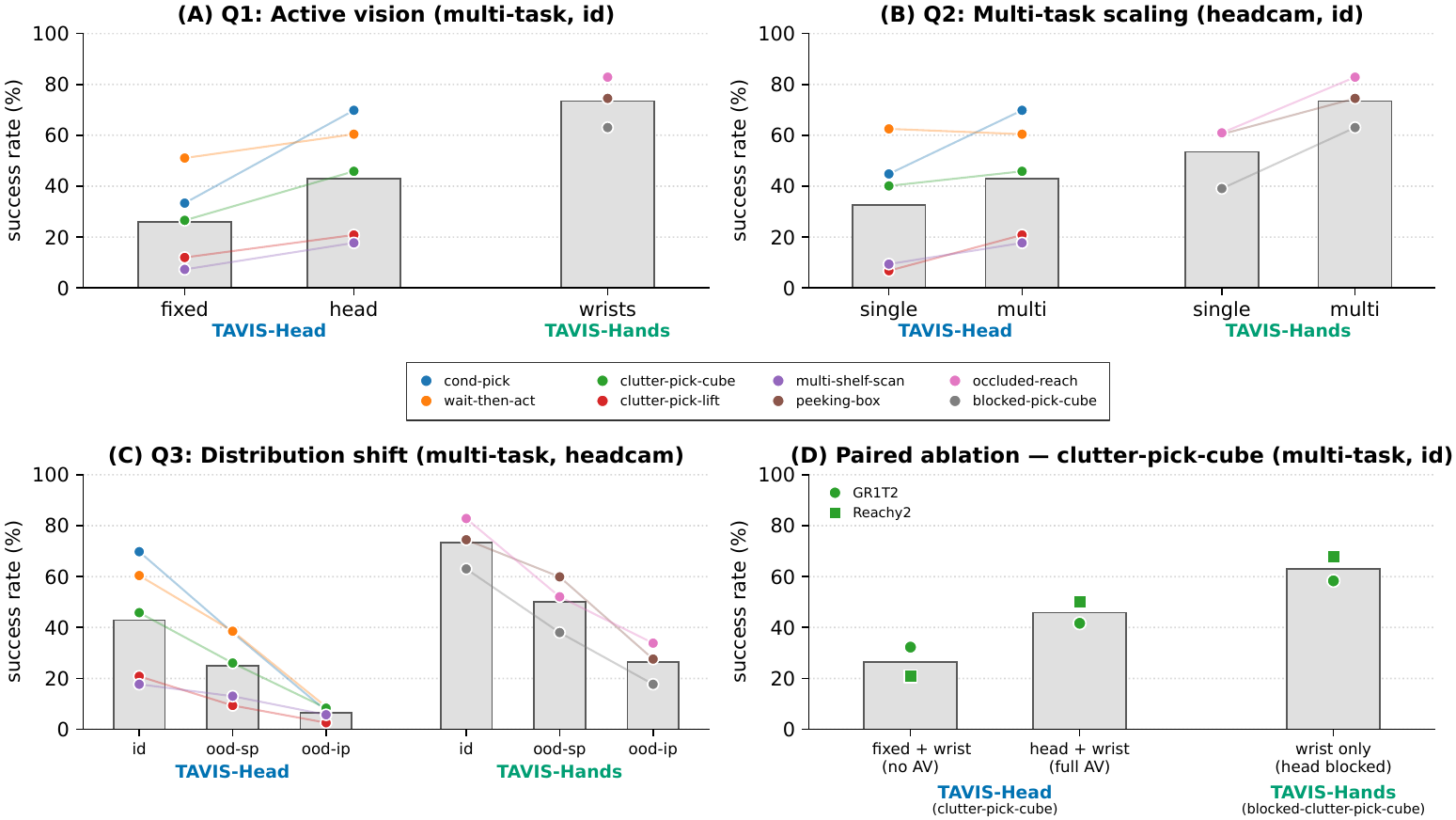}
    \caption{\textbf{TAVIS results overview.} Aggregated multi-task $\pi_0$ success rates across the four main evaluation cuts of Section~\ref{sec:results}. Bars: suite-mean SR (per-task averaged over robots); coloured dots: per-task points; thin lines: paired conditions per task. \textbf{(A)} \textit{Q1, active vision}: headcam vs fixedcam on TAVIS-Head, and head\,+\,wrist SR on TAVIS-Hands (no fixedcam variant by design). \textbf{(B)} \textit{Q2, multi-task scaling}: single-task checkpoints vs suite-multi-task $\pi_0$. \textbf{(C)} \textit{Q3, distribution shift}: id, ood-spatial (\emph{ood-sp}), ood-init-pose (\emph{ood-ip}). \textbf{(D)} \textit{Paired ablation on \textit{clutter-pick-cube}}: \textit{no AV} (TAVIS-Head fixedcam, fixed\,+\,wrist), \textit{full AV} (TAVIS-Head head\,+\,wrist), \textit{wrist only} (TAVIS-Hands \textit{blocked-clutter-pick-cube}). Per-cell numbers in Table~\ref{tab:main_results}. }

    \label{fig:results}
\end{figure}

\begin{figure}[t]
    \centering
    \includegraphics[width=\textwidth]{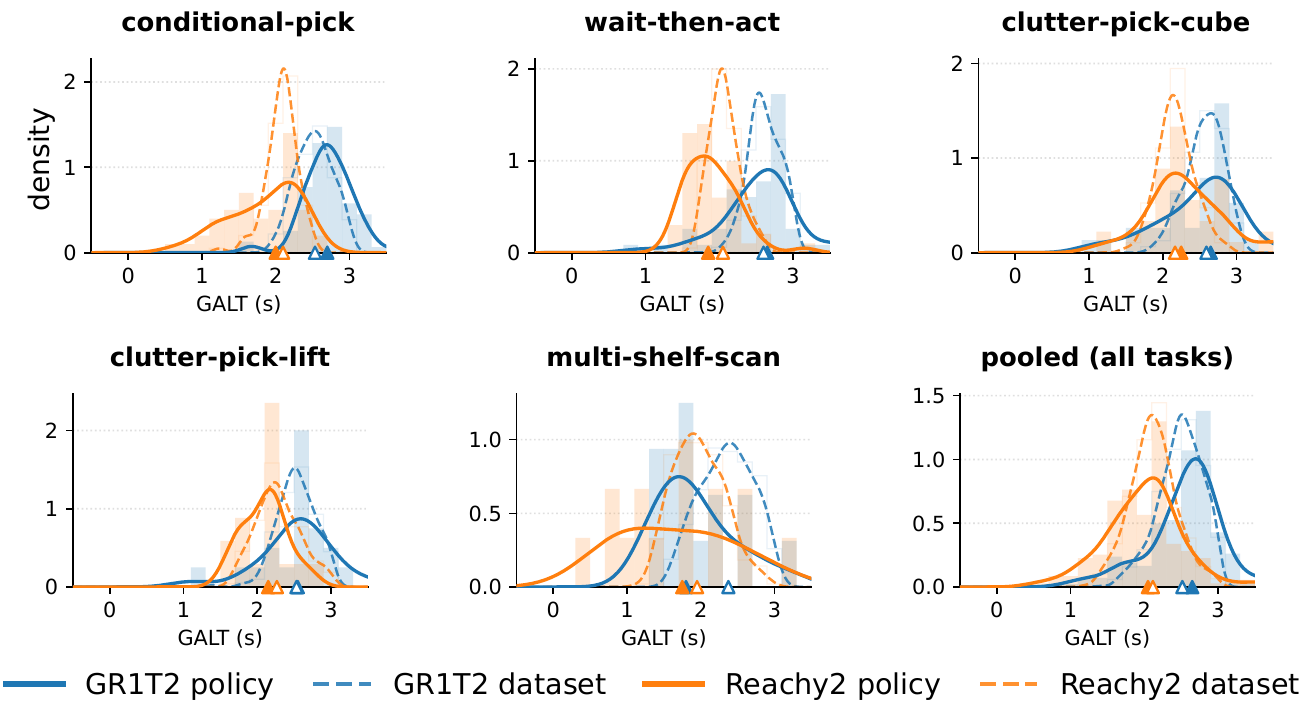}
    \caption{\textbf{GALT (Gaze-Action Lead Time) distributions per TAVIS-Head task: multi-task $\pi_0$ policy vs human demonstrations.} Solid curves are the multi-task $\pi_0$ headcam-policy GALT distribution per robot (GR1T2 blue, Reachy2 orange); dashed curves are the human teleoperation reference at the dataset's native $60$\,Hz. Light shaded histograms behind each curve use 20 equal-width bins on $[-0.5, 3.5]$\,s. Filled triangles at the $x$-axis mark policy medians, hollow triangles mark dataset medians. Only successful evaluation episodes with valid GALT detections contribute, with values outside $[-0.5, 3.0]$\,s discarded at detection time -- the same window for policy and dataset.} %
    \label{fig:galt}
\end{figure}

Full per-task results are reported in Table~\ref{tab:main_results} (multi-task $\pi_0$); single-task results for both Diffusion Policy and $\pi_0$ are in Appendix~\ref{app:suppl_results}, along with per-cell Wilson 95\% confidence intervals. Aggregated visualisations for Q1-Q3 appear in Figure~\ref{fig:results} (panels a-d), and GALT histograms for Q4 in Figure~\ref{fig:galt}. %

\begin{table}[h!]
  \centering
  \caption{\textbf{Multi-task $\pi_0$ success rates (\%) on TAVIS.} One policy per (suite, robot, camera-mode); each cell averages 96 evaluation episodes. Columns group by robot and split (\textit{id} / \textit{ood-spatial} / \textit{ood-init-pose}; defined in Section~\ref{sec:eval-splits}). For TAVIS-Head, separate headcam and fixedcam multi-task policies. For TAVIS-Hands, only the native head\,+\,wrist setup is reported (head and fixed cameras are both uninformative by design; Section~\ref{sec:eval-paired}). \textit{Suite mean} is the task average within each suite. Per-cell 95\% Wilson CIs and single-task training for $\pi_0$ and Diffusion Policy are in Appendix~\ref{app:suppl_results}. }

  \label{tab:main_results}

    {\footnotesize
    \setlength{\tabcolsep}{5pt}
    \begin{tabular}{l@{\hskip 4pt} cccccccccccc}
      \toprule
      \multirow{4}{*}{\textbf{\color[HTML]{555555} multi-task ($\pi_0$)}} & \multicolumn{6}{c}{\textit{GR1T2}} & \multicolumn{6}{c}{\textit{Reachy2}} \\
      \cmidrule(lr){2-7} \cmidrule(lr){8-13}  & \multicolumn{2}{c}{id} & \multicolumn{2}{c}{ood-spatial} & \multicolumn{2}{c}{ood-init-pose} & \multicolumn{2}{c}{id} & \multicolumn{2}{c}{ood-spatial} & \multicolumn{2}{c}{ood-init-pose}   \\
      \cmidrule(lr){2-3}\cmidrule(lr){4-5}\cmidrule(lr){6-7}\cmidrule(lr){8-9}\cmidrule(lr){10-11}\cmidrule(lr){12-13}  & head & fixed & head & fixed & head & fixed & head & fixed & head & fixed & head & fixed \\
      \midrule
      \multicolumn{13}{l}{\textbf{\color[HTML]{0072b2}TAVIS-Head}} \vspace{0.05in} \\
      conditional-pick & 87.5 & 59.4 & 49.0 & 32.3 & 2.1 & 12.5 & 52.1 & 7.3 & 27.1 & 10.4 & 12.5 & 9.4 \\
      wait-then-act & 65.6 & 88.5 & 44.8 & 61.5 & 3.1 & 15.6 & 55.2 & 13.5 & 32.3 & 5.2 & 14.6 & 10.4 \\
      clutter-pick-cube & 41.7 & 32.3 & 26.0 & 27.1 & 0.0 & 12.5 & 50.0 & 20.8 & 26.0 & 12.5 & 16.7 & 8.3 \\
      clutter-pick-lift & 22.9 & 13.5 & 9.4 & 11.5 & 0.0 & 4.2 & 18.8 & 10.4 & 9.4 & 9.4 & 5.2 & 4.2 \\
      multi-shelf-scan & 17.7 & 0.0 & 10.4 & 4.2 & 4.2 & 5.2 & 17.7 & 14.6 & 15.6 & 7.3 & 7.3 & 9.4 \\
      \midrule
      \textit{suite mean} & 47.1 & 38.8 & 27.9 & 27.3 & 1.9 & 10.0 & 38.8 & 13.3 & 22.1 & 9.0 & 11.2 & 8.3 \\
      \midrule
      \multicolumn{13}{l}{\textbf{\color[HTML]{009e73}TAVIS-Hands}} \vspace{0.05in} \\
      peeking-box & \multicolumn{2}{c}{64.6} & \multicolumn{2}{c}{51.0} & \multicolumn{2}{c}{15.6} & \multicolumn{2}{c}{84.4} & \multicolumn{2}{c}{68.8} & \multicolumn{2}{c}{39.6} \\
      occluded-reach & \multicolumn{2}{c}{87.5} & \multicolumn{2}{c}{60.4} & \multicolumn{2}{c}{24.0} & \multicolumn{2}{c}{78.1} & \multicolumn{2}{c}{43.8} & \multicolumn{2}{c}{43.8} \\
      blocked-clutter-pick-cube & \multicolumn{2}{c}{58.3} & \multicolumn{2}{c}{35.4} & \multicolumn{2}{c}{4.2} & \multicolumn{2}{c}{67.7} & \multicolumn{2}{c}{40.6} & \multicolumn{2}{c}{31.2} \\
      \midrule
      \textit{suite mean} & \multicolumn{2}{c}{70.1} & \multicolumn{2}{c}{49.0} & \multicolumn{2}{c}{14.6} & \multicolumn{2}{c}{76.7} & \multicolumn{2}{c}{51.0} & \multicolumn{2}{c}{38.2} \\
      \bottomrule
    \end{tabular}}

\end{table}

\paragraph{\textbf{Q1}: How much does active vision help, and on which task types?}

On TAVIS-Head, headcam outperforms fixedcam at the suite-mean level on both robots (paired protocol, Section~\ref{sec:eval-paired}; GR1T2: 47.1\% vs 38.8\%; Reachy2: 38.8\% vs 13.3\%; Table~\ref{tab:main_results}, Figure~\ref{fig:results}a), but the gap is conditional. The largest active-vision benefit appears on \textit{conditional-pick} (+28pp GR1T2, +45pp Reachy2), where gaze toward the cue card precedes the reach; conversely, \textit{wait-then-act} regresses on GR1T2 ($-23$pp). To gauge how much of these per-task differences reflects training-run variance, we retrained the GR1T2 policies with two additional seeds, as a small-scale estimate of inter-seed variance (Appendix~\ref{app:seeds}): the suite-level headcam advantage holds in every run ($+8$ to $+23$pp), as does the large gain on \textit{conditional-pick} ($+28$ to $+59$pp), whereas \textit{wait-then-act} changes sign across runs ($-23$ to $+37$pp), indicating no systematic effect of camera mode on that task. Reachy2's fixedcam baseline is uniformly weak (suite mean 13.3\%), inflating the apparent gap on that robot, so per-task structure is cleaner on GR1T2.

On TAVIS-Hands, both head and fixed cameras are structurally uninformative by design, so we report only head\,+\,wrist success rates and probe per-modality contributions through a paired ablation on the \textit{clutter-pick-cube} task family (Figure~\ref{fig:results}d): \textit{no AV} (TAVIS-Head fixedcam checkpoint, fixed\,+\,wrist), \textit{full AV} (TAVIS-Head headcam, head\,+\,wrist), and \textit{wrist only} (TAVIS-Hands' \textit{blocked-clutter-pick-cube} checkpoint, head masked). Going from \textit{no AV} (fixedcam) to \textit{full AV} (head\,+\,wrist) gives $+19$pp (45.9\% vs 26.6\%), while \textit{wrist only} reaches 63.0\% -- exceeding full AV due to demonstration design: TAVIS-Hands trains on explicitly exploratory wrist trajectories under occlusion. Across the suite, the multi-task policies reach 70--77\% id success, where the fixed/head views would fail by design. %

\paragraph{\textbf{Q2}: To what extent does multi-task training help versus single-task training?}

We compare single-task $\pi_0$ checkpoints against the suite-multi-task $\pi_0$ checkpoint on the headcam id split (Figure~\ref{fig:results}b; full per-task numbers in Appendix~\ref{app:suppl_results}). We find that multi-task training improves on the per-task baselines on both suites, consistent with prior IL scaling findings: suite-mean SR rises from 32.7\% to 43.0\% on TAVIS-Head and from 53.5\% to 73.4\% on TAVIS-Hands, with larger gains on the smaller-data Hands suite.

\paragraph{\textbf{Q3}: What is the impact of distribution shift during evaluation?}

In the spirit of LIBERO-Pro~\citep{zhou2025liberoPRO} and LIBERO-Plus~\citep{fei2025liberoPLUS} but on new tasks, robots, and with ray-traced rendering, we evaluate the multi-task $\pi_0$ headcam policies on the three TAVIS splits (Section~\ref{sec:eval-splits}; results in Figure~\ref{fig:results}c and Table~\ref{tab:main_results}). Performance degrades substantially under both controlled OOD shifts: TAVIS-Head suite-mean SR drops from 43.0\% (id) to 25.0\% (ood-spatial) and 6.6\% (ood-init-pose), and TAVIS-Hands from 73.4\% to 50.0\% and 26.4\% respectively. The ood-init-pose collapse on TAVIS-Head headcam exceeds that of TAVIS-Head fixedcam and TAVIS-Hands, consistent with head-pose perturbation only surfacing visually for cameras that track the head.

\paragraph{ \textbf{Q4}: Do policies acquire anticipatory gaze from imitation alone, and how does it compare to the human demonstrations? }%

For each (task, robot), we compare the policy's GALT distribution on TAVIS-Head id-split successful episodes against the dataset reference (Figure~\ref{fig:galt}), and find that multi-task $\pi_0$ headcam policies acquire anticipatory gaze comparable to that of the human demonstrations. Policy GALT leads the grasp by $\sim$2--3\,s; pooled medians agree with the dataset reference within $\sim$180\,ms on a $\sim$2.1--2.7\,s scale, and per-task $|\Delta\mathrm{median}|/\mathrm{median}_{\mathrm{dataset}}$ stays within $\sim$7--10\% on four of five tasks. The remaining outlier, \textit{multi-shelf-scan} ($|\Delta\mathrm{median}| \approx 450$\,ms, $\sim$20\% of the dataset median), likely reflects per-shelf fixation variability rather than absent GALT structure. Crucially, the temporal coupling between gaze and action -- not just the spatial trajectory -- is acquired through imitation.

The policies reproduce the timing of anticipatory gaze, but not its smoothness. At the policies' 20\,Hz control rate, the demonstrations contain about 3 head movements per episode of 34--37$^\circ$ each, whereas the $\pi_0$ policies make 12--14 smaller movements of about 12$^\circ$ on the same tasks, for a total head travel about 1.5$\times$ larger (Appendix~\ref{app:gaze-stats}). These extra movements are locked to the action chunks: 37--72\% of movement onsets fall on a single one of the 8 steps executed per policy call, in both $\pi_0$ and Diffusion Policy. The jitter is therefore a property of chunked action execution in the baseline policies, not of the demonstrations, whose head motion consists of a few large movements. Neither success rate nor GALT captures this effect. Reducing it, for instance with methods that smooth transitions between action chunks such as real-time chunking~\mbox{\citep{black2025realtime}}, is an interesting direction for future work, which the TAVIS gaze measurements make testable.

\paragraph{Wrist-camera ablation.} Both camera conditions also receive the two wrist cameras (Section~\ref{sec:eval-paired}). Retraining without them (GR1T2; multi-task $\pi_0$ and two single-task Diffusion Policy checkpoints; Appendix~\ref{app:wrist}) lowers success in both conditions but leaves the suite-level headcam/fixedcam ratio unchanged (1.22$\times$; 27.5\% vs 22.5\%), and on single-task \textit{conditional-pick} the gap widens from $+14.6$ to $+29.2$pp. Wrist views thus support manipulation in both conditions, but do not account for the headcam advantage, which remains without them.

\paragraph{Headcam teleoperation bias.} Fixedcam policies in Section~\ref{sec:results} are trained on fixed-camera observations from \emph{head}-teleoperated demonstrations, which could bias the headcam-vs-fixedcam comparison if head-teleop trajectories encode active-vision structure. To test this, we collect a matched fixedcam-only teleoperation dataset on \textit{wait-then-act} (GR1T2) and train a \emph{single-task} $\pi_0$ checkpoint. The policy trained on fixedcam-only teleoperation reaches 39.6\% id success, against 52.1\% for the standard fixedcam policy (trained on head-teleoperated trajectories) and 63.5\% for headcam. Head-teleoperated trajectories therefore do not disadvantage the fixedcam policy and may slightly benefit it, so the shared demonstrations are unlikely to inflate the headcam advantage reported in Q1.  

\section{Assumptions and Limitations}
\label{sec:limitations}

\noindent\textbf{Simulation only.} TAVIS is implemented purely in simulation on top of IsaacLab~\citep{mittal2025isaaclab}. Simulation is deliberate: it eliminates per-lab setup variability and lets any researcher reproduce the results, but comes at the price of a sim-to-real gap.

\noindent\textbf{Teleoperation setup.} We adopt by design specific and consistent fixation patterns (i.e., forcing fixations through head movement, Section~\ref{sec:dataset}). Alternative collection strategies (naive head-tracks-hands, GALT-relabelled trajectories, decoupled head/hand action lag) are interesting open directions. Lastly, the same trajectories serve both headcam and fixedcam policies, so fixedcam inherits head-teleop look-then-reach pauses; the bias investigation in Section~\ref{sec:results} finds that head-teleoperated trajectories do not disadvantage the fixedcam policy, and may slightly benefit it.  

\noindent\textbf{Restricted active-vision configurations.} TAVIS focuses on commodity 2--3-DoF pan/tilt necks (TAVIS-Head) and bimanual wrist cameras (TAVIS-Hands). The 6--7-DoF active necks explored in prior work~\citep{chuang2025active, xiong2025vision} and the eye-gimbal configurations~\citep{kerr2025eye} are not directly supported, although TAVIS's compositional design admits their addition.

\noindent\textbf{Task scope.} Each TAVIS task isolates a single gaze function within a short episode. Gaze for monitoring contact during manipulation (e.g., inspecting an insertion) and gaze allocation across multi-stage tasks are not covered; both can be added through the same task-building interface, and GALT already supports multiple action anchors per episode (Section~\ref{sec:galt}).

\noindent\textbf{GALT scope.} GALT is a \emph{post-hoc} metric on successful episodes with correct fixation. Indeed, its direct optimization (e.g., RL fine-tuning) may reward any jerky pre-grasp head motions; meaningful use requires the underlying policy to already exhibit coherent fixation. GALT also uses head orientation as a proxy for gaze direction (TAVIS robots have no independent eye DoFs), so absolute values are not directly comparable to eye-tracking literature. Also, GALT only measures timing; where the fixation lands is assessed separately, in simulation, against ground-truth geometry (Appendix~\ref{app:galt-algo}).

\noindent\textbf{OOD splits.} OOD-spatial and OOD-init-pose target only two pre-defined axes, but real-world distribution shift is richer (e.g.\ visual textures, lighting, object semantics). Notably, most tasks use a common scene template (the same metal table in an open arena). Scene-level distribution shift is not evaluated, but is a useful OOD scenario to explore. Extending TAVIS to support more OOD cases is relatively straightforward, and will be useful future work. %

\noindent\textbf{Single-seed evaluation.} All cells use a single seed per configuration, as in standard IL benchmarks~\citep{liu2023libero}. Aggregate trends are more reliable than individual cells. A small-scale estimate of inter-seed variance (Appendix~\ref{app:seeds}) supports this reading: the suite-level headcam advantage holds across three training runs while its magnitude varies ($+8$ to $+23$pp), and individual cells vary more, with \textit{wait-then-act} changing sign across runs.    

\section{Conclusion}

TAVIS provides reproducible evaluation infrastructure for egocentric active-vision imitation learning: two complementary task suites (TAVIS-Head, TAVIS-Hands), two humanoid torso embodiments under a unified canonical action space, simultaneous head- and fixed-camera recording on identical demonstrations, ID/OOD distribution splits, and GALT -- a kinematic metric that quantifies anticipatory gaze in learned policies. Our baselines show that active-vision benefits are task-conditional rather than uniform, that policies degrade sharply under controlled distribution shifts, and that imitation alone yields anticipatory gaze with median lead times comparable to those of the human demonstrations. The gaze measurements also reveal an effect that is invisible to task success: the policies' head motion is less smooth than in the demonstrations, with the extra movements concentrated at action-chunk boundaries. Evaluating active-vision policies therefore requires measuring gaze behaviour itself, not only task completion. 

The benchmark is designed to grow. Adding a new humanoid torso requires only its USD, joint indices, gripper interface, and a hip-frame offset; new tasks fit into the suite-builder interface; and the existing OOD framework extends naturally to additional perturbation axes -- visual textures, semantic substitutions, scene-level changes, and language-prompt variations are all straightforward to add given TAVIS's procedural environments. Beyond these axes, several directions are particularly promising: post-hoc gaze relabeling that retrofits anticipatory fixations onto the majority of existing IL datasets which lack head tracking; sim-to-real bridges for the established TAVIS tasks; foveated-vision variants leveraging eye-tracking, in the spirit of GIAVA~\citep{chuang2025look}; and policies whose gaze stays smooth across action chunks.

\ifanonsubmission\else
\subsubsection*{Acknowledgments}
This work used the Dutch national e-infrastructure with the support of the SURF Cooperative using grant no.\ EINF-17183. The author thanks Murat Kirtay and Bosong Ding (AIR-Lab, Tilburg University) for valuable feedback on TAVIS.

\fi

\bibliographystyle{tmlr}
\bibliography{references}

\clearpage
\newpage
\appendix

\section*{Supplementary Material}

\section{Task Specifications}
\label{appendix:tasks_info}

\noindent\textbf{Shared configuration.} TAVIS-Head tasks use a fixed pool of 5 YCB objects (\textit{soup can, meat can, tuna can, gelatin box, pudding box}; uniformly scaled $0.75\times$) on a $1.0\,$m-high table. The fixed camera is positioned at $(0.2, 0, 1.43)\,$m (thus above the workspace, slightly in front of the robot), and angled $\sim 45^\circ$ downward to provide a $120^\circ$ wide-angle coverage of the manipulation area, recording $640\times480$ RGB. Scene lighting is held constant ($800$\,lux dome + $2000$\,lux distant). Episodes are capped at $20\,$s; teleoperation episodes are operator-terminated (no enforced cap).

Per-episode variance at evaluation arises entirely from task-scene randomization; the deterministic post-reset state lies mostly inside the training start-state distribution on the axes we measured (Appendix~\ref{app:demonstrations_and_vr}), ruling out first-frame OOD as a confound. %

\noindent\textbf{Prompts.}
\begin{itemize}[leftmargin=*, noitemsep, topsep=2pt]
\item \textit{conditional-pick}: \texttt{"Look at the card. If it is red, pick the object on the left. If it is green, pick the object on the right."}
\item \textit{wait-then-act}: \texttt{"Watch the red light. When it turns green, pick up the object."}
\item \textit{clutter-pick-cube} \& \textit{blocked-clutter-pick-cube}: \texttt{"Find the red cube and pick it up."}
\item \textit{clutter-pick-lift}: object-conditional ($3$ phrasings $\times$ $5$ objects $=$ $15$ prompts; e.g., \texttt{"Pick up the tomato soup can and lift it."}).
\item \textit{multi-shelf-scan}: object-conditional ($15$ prompts; e.g., \texttt{"Find the tomato soup can on the shelf and bring it to me."}).
\item \textit{peeking-box}: \texttt{"Retrieve the object from inside the box."}
\item \textit{occluded-reach}: \texttt{"Reach around the screen and pick up the object behind it."}
\end{itemize}

\noindent\textbf{Object-conditional prompts.} \textit{clutter-pick-lift} and \textit{multi-shelf-scan} each use 3 phrasings per object across the 5 TAVIS-Head YCB objects (15 prompts per task):

\begin{itemize}[leftmargin=*, noitemsep, topsep=2pt]
\item \textit{\textbf{clutter-pick-lift}}:
\begin{itemize}[leftmargin=*, noitemsep, topsep=0pt]
\item \textit{soup can}: \texttt{"Pick up the tomato soup can and lift it."} / \texttt{"Grasp the soup can and hold it up."} / \texttt{"Lift the red soup can off the table."}
\item \textit{meat can}: \texttt{"Pick up the potted meat can and lift it."} / \texttt{"Grasp the can of spam and hold it up."} / \texttt{"Lift the meat can off the table."}
\item \textit{tuna fish can}: \texttt{"Pick up the tuna fish can and lift it."} / \texttt{"Grasp the tuna can and hold it up."} / \texttt{"Lift the tuna fish can off the table."}
\item \textit{gelatin box}: \texttt{"Pick up the gelatin box and lift it."} / \texttt{"Grasp the gelatin box and hold it up."} / \texttt{"Lift the gelatin box off the table."}
\item \textit{pudding box}: \texttt{"Pick up the pudding box and lift it."} / \texttt{"Grasp the pudding box and hold it up."} / \texttt{"Lift the pudding box off the table."}
\end{itemize}
\item \textit{\textbf{multi-shelf-scan}}:
\begin{itemize}[leftmargin=*, noitemsep, topsep=0pt]
\item \textit{soup can}: \texttt{"Find the tomato soup can on the shelf and bring it to me."} / \texttt{"Retrieve the soup can from the shelves."} / \texttt{"Look through the shelves, find the red soup can, and take it."}
\item \textit{meat can}: \texttt{"Find the potted meat can on the shelf and bring it to me."} / \texttt{"Retrieve the spam can from the shelves."} / \texttt{"Look through the shelves, find the meat can, and take it."}
\item \textit{tuna fish can}: \texttt{"Find the tuna fish can on the shelf and bring it to me."} / \texttt{"Retrieve the tuna can from the shelves."} / \texttt{"Look through the shelves, find the tuna can, and take it."}
\item \textit{gelatin box}: \texttt{"Find the gelatin box on the shelf and bring it to me."} / \texttt{"Retrieve the gelatin box from the shelves."} / \texttt{"Look through the shelves, find the gelatin box, and take it."}
\item \textit{pudding box}: \texttt{"Find the pudding box on the shelf and bring it to me."} / \texttt{"Retrieve the pudding box from the shelves."} / \texttt{"Look through the shelves, find the pudding box, and take it."}
\end{itemize}
\end{itemize}

\noindent\textbf{Per-task scenes, randomization, and success criteria.} Table~\ref{tab:task-specs} summarises the per-task scene composition, in-distribution (\textit{id}) spatial ranges, the wider \textit{ood-spatial} ranges, and the success criterion. The \textit{ood-init-pose} split applies a global Gaussian perturbation to the robot's reset pose (Section~\ref{sec:eval-splits})
and is not task-specific.

\begin{table}[h]
\centering
\caption{\textbf{Per-task specifications.} Scene composition and randomization ranges for the 8 TAVIS tasks. Position ranges are in metres; rotations in degrees. Success requires the listed object position threshold and end-effector velocity $<1$\,m/s (held briefly to avoid mid-flight detection).}
\label{tab:task-specs}

\noindent\makebox[\textwidth][c]{%
{\scriptsize
\setlength{\tabcolsep}{4pt}
\begin{tabular}{@{}lp{2.8cm}p{4.0cm}p{4.0cm}p{2.5cm}@{}}
\toprule
Task & Scene & ID area & OOD area & Success \\
\midrule
\addlinespace[2pt]
\multicolumn{5}{@{}l@{}}{\textit{TAVIS-Head}} \\[2pt]
conditional-pick & 2 YCB + cue card & objs: $10\times10$\,cm strips at $|y|=20$\,cm; card: $10\times10$\,cm centred & objs: $20\times25$\,cm strips at $|y|\approx22$\,cm; card: $20\times16$\,cm centred & target $z>1.2$\,m \\[2pt]
wait-then-act & 1 YCB + signal light & obj: $10\times24$\,cm; light: $10\times20$\,cm at $x=0.65$\,m; cue delay $\in[2,5]$\,s & obj: $20\times40$\,cm; light: $20\times30$\,cm; delay $\in[2,8]$\,s & light green AND target $z>1.2$\,m \\[2pt]
clutter-pick-cube & 4 YCB distractors + red cube ($5.5$\,cm) & $10\times50$\,cm; min separation $10$\,cm & $20\times70$\,cm; min separation $5$\,cm & cube $z>1.2$\,m \\[2pt]
clutter-pick-lift & 5 YCB; one is target & same as clutter-pick-cube & same as clutter-pick-cube & target $z>1.2$\,m \\[2pt]
multi-shelf-scan & 5 YCB on 3-shelf unit (heights $0.97/1.10/1.27$\,m) & per-slot jitter $\Delta x, \Delta y \leq 3$\,cm & $\Delta y \leq 10$\,cm, $\Delta x \leq 3$\,cm & target $x<0.46$\,m \\[2pt]
\midrule
\addlinespace[2pt] 
\multicolumn{5}{@{}l@{}}{\textit{TAVIS-Hands}} \\[2pt]
peeking-box & 1 YCB inside open-side box ($20{\times}14{\times}20$\,cm) & box $\pm(2,4)$\,cm, yaw $\pm5^\circ$; obj inside $\pm(4,2)$\,cm & box $\pm(4,4)$\,cm, yaw $\pm10^\circ$; obj inside $\pm(4,4)$\,cm & target $z>1.25$\,m \\[2pt]
occluded-reach & 1 YCB behind screen ($16{\times}40$\,cm panel at $x=0.27$\,m) & obj $10\times40$\,cm & obj $17\times60$\,cm & target $z>1.25$\,m \\[2pt]
blocked-clutter-pick-cube & inherits clutter-pick-cube; head camera blacked out & inherited from clutter-pick-cube & inherited from clutter-pick-cube & cube $z>1.2$\,m \\
\bottomrule
\end{tabular}}%
}

\end{table}

\newpage
\section{Robot Specifications}
\label{appendix:robots}

Both TAVIS robots share a unified $19$-dimensional canonical action space, abstracting away the underlying models to enable cross-embodiment evaluation:

\begin{itemize}[leftmargin=*, noitemsep, topsep=2pt]
\item \textbf{Indices 0--6}: left-arm IK target, $(x, y, z, q_w, q_x, q_y, q_z)$ in canonical
frame.
\item \textbf{Indices 7--13}: right-arm IK target, same parameterisation.
\item \textbf{Indices 14--16}: head roll, pitch, yaw (radians, absolute).
\item \textbf{Indices 17--18}: left, right gripper, normalised $[-1, 1]$ scalar.
\end{itemize}

A canonical-frame wrapper translates EEF targets between the canonical hip-centric frame and each robot's root frame using a fixed offset; orientation, head, and gripper actions pass through unchanged. Arm IK uses damped least-squares null-space solving. Per-robot models include additional locked DoFs (waist, legs, mobile base, antennas) clamped via high stiffness ($10^7$); the canonical action space exposes only the controlled joints. Grippers differ in hardware (GR1T2: Robotiq 2F-85 parallel; Reachy2: custom Pollen with mimic fingers) but expose the same scalar action. Neck roll is implemented in the library and Quest~3 teleoperation app, but disabled in our experiments.

Because of the canonical action space and composable nature of the TAVIS codebase, it is possible to extend the benchmark with new robots, only requiring the new robot's USD, joint indices, gripper interface, and the hip-frame offset.

\noindent\textbf{Cameras.} All on-board cameras (head, left wrist, right wrist) record $640\times480$ RGB. Head camera FOV $\approx 70^\circ$ (focal length $15\,$mm); wrist cameras $\approx 53^\circ$ ($21\,$mm). Camera link mounts vary slightly across robots due to head/hand-link geometry differences; the canonical role -- on-board observation streams synchronised to the agent's gaze and hand motion -- is identical.

\noindent\textbf{Control frequency.} Teleoperation is recorded at $60\,$Hz. Policies are queried at $20\,$Hz during training and evaluation (downsampled by a factor of $3$, matching common practice in the field).

\section{Demonstration Collection Protocol}
\label{app:demonstrations_and_vr}

\noindent\textbf{Hardware.} All demonstrations were collected  using a Meta Quest~3 VR headset over a wired link, with the simulated head-camera view streamed to the HMD and bimanual controllers driving the robot's arms.

\noindent\textbf{Per-episode protocol.} On reset, the robot's pose snaps to track the operator's current controller pose; the operator scans the scene briefly via head movement, returns to a near-default gaze, and then begins the recorded segment. Recording therefore starts from an operator-driven, near-canonical pose rather than the deterministic post-reset state, producing realistic mildly varied starting conditions and motivating the \textit{ood-init-pose} evaluation axis (Section~\ref{sec:eval-splits}).

\begin{figure}[h!]
    \centering
    \includegraphics[width=\textwidth]{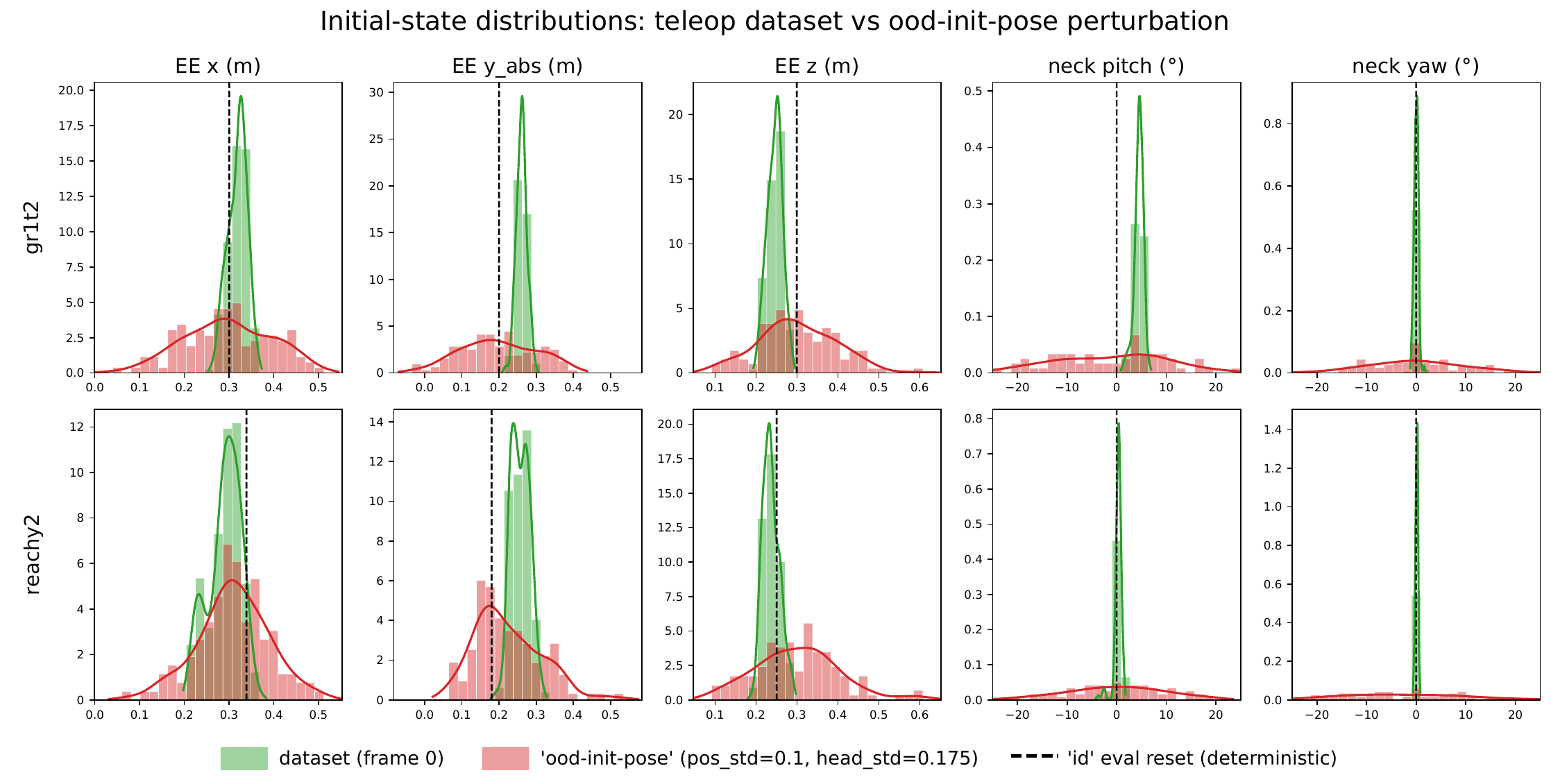}
    \caption{ \textbf{Initial-state distributions: teleop dataset, \textit{id} eval reset, and \textit{ood-init-pose} perturbation.} Rows: robot (GR1T2 top, Reachy2 bottom). Columns: end-effector position $x$, $|y|$, $z$ (metres), and neck pitch, yaw (degrees). Histograms and KDEs compare the frame-$0$ distribution in the teleoperation dataset (green) with the \textit{ood-init-pose} eval distribution (red, $\sigma_{\mathrm{pos}}=0.1$\,m and $\sigma_{\mathrm{head}}=0.175$\,rad\,$\approx 10^\circ$); the deterministic \textit{id} eval reset, a single value per dimension, is shown as a black dashed line. End-effector $y$ is reported in absolute value because demonstrations are bilaterally symmetric.}
    \label{fig:suppl-init-pose-distributions}
\end{figure}

\noindent\textbf{Initial-state distributions.} Figure~\ref{fig:suppl-init-pose-distributions} compares the teleop dataset's frame-$0$ distribution (green), the \textit{id} eval reset (black dashed), and the \textit{ood-init-pose} eval distribution (red) per robot. The \textit{ood-init-pose} perturbation extends well beyond dataset support on every dimension, especially on neck pitch and yaw where the dataset starts each episode at a near-fixed pose. Some sampled poses are physically awkward (e.g., EEF intersecting task geometry); since the same perturbation is applied uniformly across all checkpoints (Section~\ref{sec:eval-splits}), absolute success rates may be biased downward but cross-method comparisons remain valid.

\section{Supplementary Methods}
\label{app:suppl_methods}

\noindent\textbf{Baseline training setup.} Both Diffusion Policy~\citep{chi2023diffusionpolicy} and $\pi_0$~\citep{black2024pi0} are trained via their LeRobot~\citep{cadene2024lerobot} implementations. $\pi_0$ checkpoints fine-tune from the official LeRobot pretrained baseline (\texttt{lerobot/pi0\_base}) at two scopes: \textit{single-task} (one checkpoint per (suite, robot, camera, task) tuple) and \textit{multi-task} (one checkpoint per (suite, robot, camera) trained jointly over all suite tasks). Diffusion Policy is trained single-task only and only on tasks without language conditioning (\textit{clutter-pick-lift} and \textit{multi-shelf-scan} are excluded). Each setting trains two separate checkpoints -- one for headcam observations and one for fixedcam.

\noindent\textbf{Training duration.} $\pi_0$ multi-task checkpoints were trained for $120{\rm k}$ steps initially. For GR1T2 this produced strong results across all TAVIS-Head tasks. For Reachy2, the $120{\rm k}$ checkpoint underperformed the corresponding single-task baselines on several tasks (notably \textit{wait-then-act} fixedcam at ${\sim}5\%$ SR), suggesting overfitting on its smaller effective dataset. A second Reachy2 multi-task checkpoint was trained for $60{\rm k}$ steps with identical hyperparameters and seed; this checkpoint recovered SR on the previously-collapsed cells (e.g., \textit{wait-then-act} fixedcam: $5\%\!\to\!13\%$ id; headcam: $49\%\!\to\!55\%$) without regressing elsewhere. We report the $60{\rm k}$ checkpoint as canonical for Reachy2 multi-task and the $120{\rm k}$ checkpoint for GR1T2 multi-task; the choice of $60{\rm k}$ was guided by training-loss curves (still decreasing at $60{\rm k}$ for GR1T2, plateaued for Reachy2) rather than eval SR. A more careful sweep over training duration is left to future work.

\noindent\textbf{Constant-dimension normalization.} A subtle failure mode arises in sim-based IL when observation dimensions have near-zero variance in the training data (e.g., locked joints, passive DoFs left to settle under gravity): standard mean/std normalization produces large coefficients that amplify tiny eval-time deviations into catastrophic apparent OOD signals. We implement a preprocessing step (\texttt{fix\_constant\_dims}) that detects these dimensions and applies identity normalization to them. This affected Reachy2 (where settle dynamics on its mobile base differ from GR1T2's grounded torso) and was not necessary for GR1T2.

\noindent\textbf{Confidence intervals.} All success-rate confidence intervals reported in this paper use the Wilson score interval at $\alpha\!=\!0.05$, computed via \texttt{statsmodels.stats.proportion.proportion\_confint} with \texttt{method=`wilson'}.

\noindent\textbf{Hyperparameters and compute.} Training hyperparameters and wall-clock estimates are listed in Table~\ref{tab:hparams}. Individual training runs used single NVIDIA H100 GPUs, while evaluation was run on a single NVIDIA GeForce RTX 4090, since raytracing is required.

\begin{table}[h]
\centering
\caption{\textbf{Training hyperparameters.} Both baselines use LeRobot defaults except where
noted.}
\label{tab:hparams}
\small
\renewcommand{\arraystretch}{1.1}
\begin{tabular}{@{}lcc@{}}
\toprule
& Diffusion Policy & $\pi_0$ \\
\midrule
Parameters & $272.5$\,M & $3.5$\,B \\
Optimizer & AdamW & AdamW \\
Learning rate & $1\!\times\!10^{-4}$ & $2.5\!\times\!10^{-5}$ \\
Weight decay & $1\!\times\!10^{-6}$ & $1\!\times\!10^{-2}$ \\
Batch size (per device) & $16$ & $16$ \\
Mixed precision & bfloat16 & bfloat16 \\
Observation horizon & $2$ frames & $1$ frame \\
Prediction horizon & $16$ frames (0.8s) & $16$ frames (0.8s) \\
Action chunk size & $8$ frames (0.4s) & $8$ frames (0.4s) \\
Image resolution & $320 \times 240$ & $224 \times 224$ (PaLiGemma) \\
Noise schedule & DDPM, sq.cos.cap.v2 ($100$ steps) & --- \\
Training steps (single-task) & $200{\rm k}$ & $15{\rm k}$ \\
Training steps (multi-task) & --- & $60{\rm k}$ (Reachy2) / $120{\rm k}$ (GR1T2) \\
\bottomrule
\end{tabular}
\end{table}

\section{Supplementary Results}
\label{app:suppl_results}

\subsection{Full Results}
\label{app:per_cell}

We report per-cell results that complement the main-text Table~\ref{tab:main_results}. Three tables follow, all using the same $96$-episode-per-cell evaluation protocol and $95\%$ Wilson confidence intervals: multi-task $\pi_0$ with explicit CIs (Table~\ref{tab:main_results_w_ci}, mirroring Table~\ref{tab:main_results}), single-task $\pi_0$ checkpoints across all (suite, robot, camera, task) tuples (Table~\ref{tab:main_results_singletask_pi0}), and single-task Diffusion Policy checkpoints (Table~\ref{tab:main_results_singletask_diffpol}). Single-task DP entries for the two language-conditioned TAVIS-Head tasks (\textit{clutter-pick-lift}, \textit{multi-shelf-scan}) are absent by design.

\noindent\textbf{Diffusion Policy vs $\pi_0$ (single-task).} The two methods produce qualitatively similar findings on TAVIS: headcam-vs-fixedcam gaps on TAVIS-Head, OOD degradation patterns, and per-task ordering are largely consistent. In absolute single-task ID success, Diffusion Policy is competitive with -- and on several TAVIS-Head tasks, outperforms -- $\pi_0$ at the same scope (e.g., GR1T2 headcam suite mean: $54.2\%$ vs $25.8\%$). This is not an apples-to-apples comparison: $\pi_0$ single-task fine-tunes a strong pretrained checkpoint for $15{\rm k}$ steps, while Diffusion Policy trains from scratch for $200{\rm k}$ steps. $\pi_0$'s relative advantage materialises in the multi-task scope (Table~\ref{tab:main_results_w_ci}), which Diffusion Policy cannot access in its standard form due to the absence of language conditioning.

\noindent\textbf{Implicit language-prompting comparison.} Two of the five TAVIS-Head tasks -- \textit{clutter-pick-lift} and \textit{multi-shelf-scan} -- include natural-language conditioning across three prompt variants for each object in the task (i.e., 15 different prompts per task), while the other three use a single fixed objective. Despite 250 vs 100 demonstrations, the language-prompted tasks reach only $\sim$19\% multi-task $\pi_0$ id headcam SR (averaged across robots) against $\sim$59\% for the non-prompted ones. We do not read this as a clean language-vs-no-language ablation: the language tasks are also our most complex in absolute terms, and per (object, prompt) tuple they actually receive fewer demonstrations than the non-prompted counterparts ($\sim$17 vs 20). Disentangling language conditioning from task complexity at matched data budgets is left to future work; we report the gap here as an implicit observation built into the existing benchmark design.

\begin{table}[h!]
  \centering
  \caption{\textbf{Multi-task $\pi_0$ success rates (\%) on TAVIS, with 95\% Wilson confidence intervals.} Same checkpoints, evaluation episodes (96 per cell), column structure, and split definitions as Table~\ref{tab:main_results}; intervals are computed per cell using the Wilson score method.}
  \label{tab:main_results_w_ci}

    \noindent\makebox[\textwidth][c]{%
    {\scriptsize
    \setlength{\tabcolsep}{2.6pt}\renewcommand{\arraystretch}{1.1}
    \begin{tabular}{l@{\hskip 4pt} cccccccccccc}
      \toprule
      \multirow{4}{*}{\textbf{\color[HTML]{555555} multi-task ($\pi_0$, 95\% CI)}} & \multicolumn{6}{c}{\textit{GR1T2}} & \multicolumn{6}{c}{\textit{Reachy2}} \\
      \cmidrule(lr){2-7} \cmidrule(lr){8-13}  & \multicolumn{2}{c}{id} & \multicolumn{2}{c}{ood-spatial} & \multicolumn{2}{c}{ood-init-pose} & \multicolumn{2}{c}{id} & \multicolumn{2}{c}{ood-spatial} & \multicolumn{2}{c}{ood-init-pose}   \\
      \cmidrule(lr){2-3}\cmidrule(lr){4-5}\cmidrule(lr){6-7}\cmidrule(lr){8-9}\cmidrule(lr){10-11}\cmidrule(lr){12-13}  & head & fixed & head & fixed & head & fixed & head & fixed & head & fixed & head & fixed \\
      \midrule
      \multicolumn{13}{l}{\textbf{\color[HTML]{0072b2}TAVIS-Head}} \vspace{0.05in} \\
      conditional-pick & \srci{87.5}{79.4,92.7} & \srci{59.4}{49.4,68.7} & \srci{49.0}{39.2,58.8} & \srci{32.3}{23.8,42.2} & \srci{2.1}{0.6,7.3} & \srci{12.5}{7.3,20.6} & \srci{52.1}{42.2,61.8} & \srci{7.3}{3.6,14.3} & \srci{27.1}{19.2,36.7} & \srci{10.4}{5.8,18.1} & \srci{12.5}{7.3,20.6} & \srci{9.4}{5.0,16.9} \\\addlinespace[1.5pt]
      wait-then-act & \srci{65.6}{55.7,74.4} & \srci{88.5}{80.6,93.5} & \srci{44.8}{35.2,54.7} & \srci{61.5}{51.5,70.6} & \srci{3.1}{1.1,8.8} & \srci{15.6}{9.7,24.2} & \srci{55.2}{45.3,64.8} & \srci{13.5}{8.1,21.8} & \srci{32.3}{23.8,42.2} & \srci{5.2}{2.2,11.6} & \srci{14.6}{8.9,23.0} & \srci{10.4}{5.8,18.1} \\\addlinespace[1.5pt]
      clutter-pick-cube & \srci{41.7}{32.3,51.7} & \srci{32.3}{23.8,42.2} & \srci{26.0}{18.3,35.6} & \srci{27.1}{19.2,36.7} & \srci{0.0}{0.0,3.8} & \srci{12.5}{7.3,20.6} & \srci{50.0}{40.2,59.8} & \srci{20.8}{13.9,30.0} & \srci{26.0}{18.3,35.6} & \srci{12.5}{7.3,20.6} & \srci{16.7}{10.5,25.4} & \srci{8.3}{4.3,15.6} \\\addlinespace[1.5pt]
      clutter-pick-lift & \srci{22.9}{15.6,32.3} & \srci{13.5}{8.1,21.8} & \srci{9.4}{5.0,16.9} & \srci{11.5}{6.5,19.4} & \srci{0.0}{0.0,3.8} & \srci{4.2}{1.6,10.2} & \srci{18.8}{12.2,27.7} & \srci{10.4}{5.8,18.1} & \srci{9.4}{5.0,16.9} & \srci{9.4}{5.0,16.9} & \srci{5.2}{2.2,11.6} & \srci{4.2}{1.6,10.2} \\\addlinespace[1.5pt]
      multi-shelf-scan & \srci{17.7}{11.4,26.5} & \srci{0.0}{0.0,3.8} & \srci{10.4}{5.8,18.1} & \srci{4.2}{1.6,10.2} & \srci{4.2}{1.6,10.2} & \srci{5.2}{2.2,11.6} & \srci{17.7}{11.4,26.5} & \srci{14.6}{8.9,23.0} & \srci{15.6}{9.7,24.2} & \srci{7.3}{3.6,14.3} & \srci{7.3}{3.6,14.3} & \srci{9.4}{5.0,16.9} \\\addlinespace[1.5pt]
      \midrule
      \textit{suite mean} & \srci{47.1}{42.7,51.6} & \srci{38.8}{34.5,43.2} & \srci{27.9}{24.1,32.1} & \srci{27.3}{23.5,31.4} & \srci{1.9}{1.0,3.5} & \srci{10.0}{7.6,13.0} & \srci{38.8}{34.5,43.2} & \srci{13.3}{10.6,16.7} & \srci{22.1}{18.6,26.0} & \srci{9.0}{6.7,11.8} & \srci{11.2}{8.7,14.4} & \srci{8.3}{6.2,11.1} \\\addlinespace[1.5pt]
      \midrule
      \multicolumn{13}{l}{\textbf{\color[HTML]{009e73}TAVIS-Hands}} \vspace{0.05in} \\
      peeking-box & \multicolumn{2}{c}{\srci{64.6}{54.6,73.4}} & \multicolumn{2}{c}{\srci{51.0}{41.2,60.8}} & \multicolumn{2}{c}{\srci{15.6}{9.7,24.2}} & \multicolumn{2}{c}{\srci{84.4}{75.8,90.3}} & \multicolumn{2}{c}{\srci{68.8}{58.9,77.1}} & \multicolumn{2}{c}{\srci{39.6}{30.4,49.6}} \\\addlinespace[1.5pt]
      occluded-reach & \multicolumn{2}{c}{\srci{87.5}{79.4,92.7}} & \multicolumn{2}{c}{\srci{60.4}{50.4,69.6}} & \multicolumn{2}{c}{\srci{24.0}{16.5,33.4}} & \multicolumn{2}{c}{\srci{78.1}{68.9,85.2}} & \multicolumn{2}{c}{\srci{43.8}{34.3,53.7}} & \multicolumn{2}{c}{\srci{43.8}{34.3,53.7}} \\\addlinespace[1.5pt]
      blocked-clutter-pick-cube & \multicolumn{2}{c}{\srci{58.3}{48.3,67.7}} & \multicolumn{2}{c}{\srci{35.4}{26.6,45.4}} & \multicolumn{2}{c}{\srci{4.2}{1.6,10.2}} & \multicolumn{2}{c}{\srci{67.7}{57.8,76.2}} & \multicolumn{2}{c}{\srci{40.6}{31.3,50.6}} & \multicolumn{2}{c}{\srci{31.2}{22.9,41.1}} \\\addlinespace[1.5pt]
      \midrule
      \textit{suite mean} & \multicolumn{2}{c}{\srci{70.1}{64.6,75.1}} & \multicolumn{2}{c}{\srci{49.0}{43.2,54.7}} & \multicolumn{2}{c}{\srci{14.6}{11.0,19.1}} & \multicolumn{2}{c}{\srci{76.7}{71.5,81.2}} & \multicolumn{2}{c}{\srci{51.0}{45.3,56.8}} & \multicolumn{2}{c}{\srci{38.2}{32.8,43.9}} \\\addlinespace[1.5pt]
      \bottomrule
    \end{tabular}}%
    }

\end{table}

\begin{table}[p]
  \centering
  \caption{\textbf{Single-task $\pi_0$ success rates (\%) on TAVIS, with 95\% Wilson confidence intervals.} Each cell corresponds to an independent $\pi_0$ checkpoint trained on a single (suite, robot, camera-mode, task) tuple and evaluated for 96 episodes. Column structure and split definitions are identical to Table~\ref{tab:main_results}.}
  \label{tab:main_results_singletask_pi0}

    \noindent\makebox[\textwidth][c]{%
    {\scriptsize
    \setlength{\tabcolsep}{2.6pt}\renewcommand{\arraystretch}{1.1}
    \begin{tabular}{l@{\hskip 4pt} cccccccccccc}
      \toprule
      \multirow{4}{*}{\textbf{\color[HTML]{555555} single-task ($\pi_0$, 95\% CI)}} & \multicolumn{6}{c}{\textit{GR1T2}} & \multicolumn{6}{c}{\textit{Reachy2}} \\
      \cmidrule(lr){2-7} \cmidrule(lr){8-13}  & \multicolumn{2}{c}{id} & \multicolumn{2}{c}{ood-spatial} & \multicolumn{2}{c}{ood-init-pose} & \multicolumn{2}{c}{id} & \multicolumn{2}{c}{ood-spatial} & \multicolumn{2}{c}{ood-init-pose}   \\
      \cmidrule(lr){2-3}\cmidrule(lr){4-5}\cmidrule(lr){6-7}\cmidrule(lr){8-9}\cmidrule(lr){10-11}\cmidrule(lr){12-13}  & head & fixed & head & fixed & head & fixed & head & fixed & head & fixed & head & fixed \\
      \midrule
      \multicolumn{13}{l}{\textbf{\color[HTML]{0072b2}TAVIS-Head}} \vspace{0.05in} \\
      conditional-pick & \srci{42.7}{33.3,52.7} & \srci{59.4}{49.4,68.7} & \srci{12.5}{7.3,20.6} & \srci{19.8}{13.1,28.9} & \srci{1.0}{0.2,5.7} & \srci{10.4}{5.8,18.1} & \srci{46.9}{37.2,56.8} & \srci{43.8}{34.3,53.7} & \srci{22.9}{15.6,32.3} & \srci{11.5}{6.5,19.4} & \srci{37.5}{28.5,47.5} & \srci{37.5}{28.5,47.5} \\\addlinespace[1.5pt]
      wait-then-act & \srci{63.5}{53.6,72.5} & \srci{52.1}{42.2,61.8} & \srci{17.7}{11.4,26.5} & \srci{29.2}{21.0,38.9} & \srci{31.2}{22.9,41.1} & \srci{22.9}{15.6,32.3} & \srci{61.5}{51.5,70.6} & \srci{60.4}{50.4,69.6} & \srci{42.7}{33.3,52.7} & \srci{26.0}{18.3,35.6} & \srci{43.8}{34.3,53.7} & \srci{37.5}{28.5,47.5} \\\addlinespace[1.5pt]
      clutter-pick-cube & \srci{17.7}{11.4,26.5} & \srci{27.1}{19.2,36.7} & \srci{16.7}{10.5,25.4} & \srci{14.6}{8.9,23.0} & \srci{5.2}{2.2,11.6} & \srci{12.5}{7.3,20.6} & \srci{62.5}{52.5,71.5} & \srci{14.6}{8.9,23.0} & \srci{38.5}{29.4,48.5} & \srci{3.1}{1.1,8.8} & \srci{18.8}{12.2,27.7} & \srci{7.3}{3.6,14.3} \\\addlinespace[1.5pt]
      clutter-pick-lift & \srci{0.0}{0.0,3.8} & \srci{10.4}{5.8,18.1} & \srci{0.0}{0.0,3.8} & \srci{1.0}{0.2,5.7} & \srci{0.0}{0.0,3.8} & \srci{4.2}{1.6,10.2} & \srci{13.5}{8.1,21.8} & \srci{14.6}{8.9,23.0} & \srci{7.3}{3.6,14.3} & \srci{5.2}{2.2,11.6} & \srci{6.2}{2.9,13.0} & \srci{15.6}{9.7,24.2} \\\addlinespace[1.5pt]
      multi-shelf-scan & \srci{5.2}{2.2,11.6} & \srci{0.0}{0.0,3.8} & \srci{1.0}{0.2,5.7} & \srci{1.0}{0.2,5.7} & \srci{6.2}{2.9,13.0} & \srci{6.2}{2.9,13.0} & \srci{13.5}{8.1,21.8} & \srci{5.2}{2.2,11.6} & \srci{9.4}{5.0,16.9} & \srci{6.2}{2.9,13.0} & \srci{8.3}{4.3,15.6} & \srci{2.1}{0.6,7.3} \\\addlinespace[1.5pt]
      \midrule
      \textit{suite mean} & \srci{25.8}{22.1,29.9} & \srci{29.8}{25.9,34.0} & \srci{9.6}{7.3,12.5} & \srci{13.1}{10.4,16.4} & \srci{8.8}{6.5,11.6} & \srci{11.2}{8.7,14.4} & \srci{39.6}{35.3,44.0} & \srci{27.7}{23.9,31.9} & \srci{24.2}{20.6,28.2} & \srci{10.4}{8.0,13.5} & \srci{22.9}{19.4,26.9} & \srci{20.0}{16.7,23.8} \\\addlinespace[1.5pt]
      \midrule
      \multicolumn{13}{l}{\textbf{\color[HTML]{009e73}TAVIS-Hands}} \vspace{0.05in} \\
      peeking-box & \multicolumn{2}{c}{\srci{47.9}{38.2,57.8}} & \multicolumn{2}{c}{\srci{38.5}{29.4,48.5}} & \multicolumn{2}{c}{\srci{21.9}{14.8,31.1}} & \multicolumn{2}{c}{\srci{72.9}{63.3,80.8}} & \multicolumn{2}{c}{\srci{46.9}{37.2,56.8}} & \multicolumn{2}{c}{\srci{56.2}{46.3,65.7}} \\\addlinespace[1.5pt]
      occluded-reach & \multicolumn{2}{c}{\srci{68.8}{58.9,77.1}} & \multicolumn{2}{c}{\srci{36.5}{27.5,46.4}} & \multicolumn{2}{c}{\srci{11.5}{6.5,19.4}} & \multicolumn{2}{c}{\srci{53.1}{43.2,62.8}} & \multicolumn{2}{c}{\srci{29.2}{21.0,38.9}} & \multicolumn{2}{c}{\srci{40.6}{31.3,50.6}} \\\addlinespace[1.5pt]
      blocked-clutter-pick-cube & \multicolumn{2}{c}{\srci{44.8}{35.2,54.7}} & \multicolumn{2}{c}{\srci{21.9}{14.8,31.1}} & \multicolumn{2}{c}{\srci{10.4}{5.8,18.1}} & \multicolumn{2}{c}{\srci{33.3}{24.7,43.2}} & \multicolumn{2}{c}{\srci{32.3}{23.8,42.2}} & \multicolumn{2}{c}{\srci{25.0}{17.4,34.5}} \\\addlinespace[1.5pt]
      \midrule
      \textit{suite mean} & \multicolumn{2}{c}{\srci{53.8}{48.0,59.5}} & \multicolumn{2}{c}{\srci{32.3}{27.2,37.9}} & \multicolumn{2}{c}{\srci{14.6}{11.0,19.1}} & \multicolumn{2}{c}{\srci{53.1}{47.4,58.8}} & \multicolumn{2}{c}{\srci{36.1}{30.8,41.8}} & \multicolumn{2}{c}{\srci{40.6}{35.1,46.4}} \\\addlinespace[1.5pt]
      \bottomrule
    \end{tabular}}%
    }

\vspace{0.2in}

  \centering
  \caption{\textbf{Single-task Diffusion Policy success rates (\%) on TAVIS, with 95\% Wilson confidence intervals.} Each cell corresponds to an independent Diffusion Policy~\citep{chi2023diffusionpolicy} checkpoint trained on a single (suite, robot, camera-mode, task) tuple and evaluated for 96 episodes. Diffusion Policy is trained only on tasks without language conditioning, so \textit{clutter-pick-lift} and \textit{multi-shelf-scan} are excluded (marked `-'). Column structure and split definitions are identical to Table~\ref{tab:main_results}.}
  \label{tab:main_results_singletask_diffpol}

    \noindent\makebox[\textwidth][c]{%
    {\scriptsize
    \setlength{\tabcolsep}{2.6pt}\renewcommand{\arraystretch}{1.1}
    \begin{tabular}{l@{\hskip 4pt} cccccccccccc}
      \toprule
      \multirow{4}{*}{\textbf{\color[HTML]{555555} single-task (diffusion, 95\% CI)}} & \multicolumn{6}{c}{\textit{GR1T2}} & \multicolumn{6}{c}{\textit{Reachy2}} \\
      \cmidrule(lr){2-7} \cmidrule(lr){8-13}  & \multicolumn{2}{c}{id} & \multicolumn{2}{c}{ood-spatial} & \multicolumn{2}{c}{ood-init-pose} & \multicolumn{2}{c}{id} & \multicolumn{2}{c}{ood-spatial} & \multicolumn{2}{c}{ood-init-pose}   \\
      \cmidrule(lr){2-3}\cmidrule(lr){4-5}\cmidrule(lr){6-7}\cmidrule(lr){8-9}\cmidrule(lr){10-11}\cmidrule(lr){12-13}  & head & fixed & head & fixed & head & fixed & head & fixed & head & fixed & head & fixed \\
      \midrule
      \multicolumn{13}{l}{\textbf{\color[HTML]{0072b2}TAVIS-Head}} \vspace{0.05in} \\
      conditional-pick & \srci{47.9}{38.2,57.8} & \srci{33.3}{24.7,43.2} & \srci{21.9}{14.8,31.1} & \srci{9.4}{5.0,16.9} & \srci{12.5}{7.3,20.6} & \srci{19.8}{13.1,28.9} & \srci{40.6}{31.3,50.6} & \srci{40.6}{31.3,50.6} & \srci{21.9}{14.8,31.1} & \srci{7.3}{3.6,14.3} & \srci{21.9}{14.8,31.1} & \srci{20.8}{13.9,30.0} \\\addlinespace[1.5pt]
      wait-then-act & \srci{71.9}{62.2,79.9} & \srci{43.8}{34.3,53.7} & \srci{36.5}{27.5,46.4} & \srci{12.5}{7.3,20.6} & \srci{40.6}{31.3,50.6} & \srci{16.7}{10.5,25.4} & \srci{68.8}{58.9,77.1} & \srci{25.0}{17.4,34.5} & \srci{30.2}{21.9,40.0} & \srci{10.4}{5.8,18.1} & \srci{20.8}{13.9,30.0} & \srci{28.1}{20.1,37.8} \\\addlinespace[1.5pt]
      clutter-pick-cube & \srci{42.7}{33.3,52.7} & \srci{30.2}{21.9,40.0} & \srci{41.7}{32.3,51.7} & \srci{20.8}{13.9,30.0} & \srci{18.8}{12.2,27.7} & \srci{13.5}{8.1,21.8} & \srci{31.2}{22.9,41.1} & \srci{14.6}{8.9,23.0} & \srci{27.1}{19.2,36.7} & \srci{14.6}{8.9,23.0} & \srci{24.0}{16.5,33.4} & \srci{9.4}{5.0,16.9} \\\addlinespace[1.5pt]
      clutter-pick-lift & - & - & - & - & - & - & - & - & - & - & - & - \\
      multi-shelf-scan & - & - & - & - & - & - & - & - & - & - & - & - \\
      \midrule
      \textit{suite mean} & \srci{54.2}{48.4,59.8} & \srci{35.8}{30.4,41.5} & \srci{33.3}{28.1,39.0} & \srci{14.2}{10.7,18.7} & \srci{24.0}{19.4,29.2} & \srci{16.7}{12.8,21.4} & \srci{46.9}{41.2,52.6} & \srci{26.7}{22.0,32.1} & \srci{26.4}{21.6,31.8} & \srci{10.8}{7.7,14.9} & \srci{22.2}{17.8,27.4} & \srci{19.4}{15.3,24.4} \\\addlinespace[1.5pt]
      \midrule
      \multicolumn{13}{l}{\textbf{\color[HTML]{009e73}TAVIS-Hands}} \vspace{0.05in} \\
      peeking-box & \multicolumn{2}{c}{\srci{69.8}{60.0,78.1}} & \multicolumn{2}{c}{\srci{54.2}{44.2,63.8}} & \multicolumn{2}{c}{\srci{43.8}{34.3,53.7}} & \multicolumn{2}{c}{\srci{56.2}{46.3,65.7}} & \multicolumn{2}{c}{\srci{47.9}{38.2,57.8}} & \multicolumn{2}{c}{\srci{43.8}{34.3,53.7}} \\\addlinespace[1.5pt]
      occluded-reach & \multicolumn{2}{c}{\srci{83.3}{74.6,89.5}} & \multicolumn{2}{c}{\srci{52.1}{42.2,61.8}} & \multicolumn{2}{c}{\srci{27.1}{19.2,36.7}} & \multicolumn{2}{c}{\srci{39.6}{30.4,49.6}} & \multicolumn{2}{c}{\srci{31.2}{22.9,41.1}} & \multicolumn{2}{c}{\srci{46.9}{37.2,56.8}} \\\addlinespace[1.5pt]
      blocked-clutter-pick-cube & \multicolumn{2}{c}{\srci{37.5}{28.5,47.5}} & \multicolumn{2}{c}{\srci{28.1}{20.1,37.8}} & \multicolumn{2}{c}{\srci{17.7}{11.4,26.5}} & \multicolumn{2}{c}{\srci{28.1}{20.1,37.8}} & \multicolumn{2}{c}{\srci{22.9}{15.6,32.3}} & \multicolumn{2}{c}{\srci{15.6}{9.7,24.2}} \\\addlinespace[1.5pt]
      \midrule
      \textit{suite mean} & \multicolumn{2}{c}{\srci{63.5}{57.8,68.9}} & \multicolumn{2}{c}{\srci{44.8}{39.2,50.6}} & \multicolumn{2}{c}{\srci{29.5}{24.5,35.0}} & \multicolumn{2}{c}{\srci{41.3}{35.8,47.1}} & \multicolumn{2}{c}{\srci{34.0}{28.8,39.7}} & \multicolumn{2}{c}{\srci{35.4}{30.1,41.1}} \\\addlinespace[1.5pt]
      \bottomrule
    \end{tabular}}%
    }

\end{table}

\clearpage
\subsection{Inter-Seed Variability}
\label{app:seeds}

The main results use a single training run per configuration, as retraining every (suite, robot, camera, method, task) combination several times is computationally prohibitive. To help interpret the differences reported there, we estimate inter-seed variance on a small scale by retraining the multi-task $\pi_0$ TAVIS-Head policies on GR1T2 (headcam and fixedcam) with two additional random seeds, and evaluating them on the \textit{id} split with the same protocol as the main results (96 episodes per task). Run 1 is the run in the main results (Table~\ref{tab:main_results}), trained without an explicit random seed; runs 2 and 3 use seeds 123 and 7, with all other settings unchanged. Tables~\ref{tab:seeds_sr} and~\ref{tab:seeds_galt} report success rates and GALT for the three runs.

\noindent\textbf{Headcam vs fixedcam.} Headcam outperforms fixedcam at suite level in all three runs, by $+8.3$, $+21.9$ and $+22.9$\,pp (mean $+17.7 \pm 8.1$\,pp). The sign of the effect is therefore stable across seeds, while its magnitude varies, mostly because of the fixedcam policies (suite mean 23.8--38.8\%, against 46.7--54.6\% for headcam); the run in the main results shows the smallest gap of the three. Per task, the difference is positive in all three runs on four of the five tasks, and largest on \textit{conditional-pick} ($+28.1$ to $+59.4$\,pp).

\noindent\textbf{\textit{wait-then-act}.} We found that \textit{wait-then-act} is the only task without a consistent sign across runs: the difference is $-22.9$, $+3.1$ and $+36.5$\,pp (mean $+5.6 \pm 29.8$\,pp), and the task shows the largest run-to-run variability in both camera modes (headcam 65.6--92.7\%, fixedcam 51.0--89.6\%). The $-22.9$\,pp difference in the main results (Table~\ref{tab:main_results}) is therefore specific to that training run rather than a property of the task. This is consistent with the information structure of \textit{wait-then-act}: the relevant cue is a temporal state change that is equally legible from either viewpoint, so re-orienting the head provides no systematic advantage.

\noindent\textbf{GALT.} Anticipatory gaze is stable across runs: the pooled median GALT is 2.65, 2.50 and 2.60\,s, and per-task medians differ across runs by at most 0.3\,s, except on \textit{multi-shelf-scan} (0.4\,s, with 8--16 valid episodes per run).

\begin{table}[h!]
  \centering
  \caption{\textbf{Multi-task $\pi_0$ success rates (\%) across three independent training runs.} TAVIS-Head, GR1T2, \textit{id} split, 96 episodes per cell, with 95\% Wilson confidence intervals. Run 1 is the run in the main results (Table~\ref{tab:main_results}); runs 2 and 3 use seeds 123 and 7. $\Delta$: headcam minus fixedcam (pp).}
  \label{tab:seeds_sr}

    \noindent\makebox[\textwidth][c]{%
    {\scriptsize
    \setlength{\tabcolsep}{2.6pt}\renewcommand{\arraystretch}{1.1}
    \begin{tabular}{l@{\hskip 4pt} ccc ccc ccc c}
      \toprule
      \multirow{2}{*}{\textbf{\color[HTML]{555555} multi-task ($\pi_0$, 95\% CI)}} & \multicolumn{3}{c}{run 1 (Table~\ref{tab:main_results})} & \multicolumn{3}{c}{run 2 (seed 123)} & \multicolumn{3}{c}{run 3 (seed 7)} & across runs \\
      \cmidrule(lr){2-4}\cmidrule(lr){5-7}\cmidrule(lr){8-10}\cmidrule(lr){11-11} & head & fixed & $\Delta$ & head & fixed & $\Delta$ & head & fixed & $\Delta$ & $\Delta$ mean $\pm$ sd \\
      \midrule
      conditional-pick & \srci{87.5}{79.4,92.7} & \srci{59.4}{49.4,68.7} & $+28.1$ & \srci{83.3}{74.6,89.5} & \srci{24.0}{16.5,33.4} & $+59.4$ & \srci{67.7}{57.8,76.2} & \srci{26.0}{18.3,35.6} & $+41.7$ & $+43.1 \pm 15.7$ \\\addlinespace[1.5pt]
      wait-then-act & \srci{65.6}{55.7,74.4} & \srci{88.5}{80.6,93.5} & $-22.9$ & \srci{92.7}{85.7,96.4} & \srci{89.6}{81.9,94.2} & $+3.1$ & \srci{87.5}{79.4,92.7} & \srci{51.0}{41.2,60.8} & $+36.5$ & $+5.6 \pm 29.8$ \\\addlinespace[1.5pt]
      clutter-pick-cube & \srci{41.7}{32.3,51.7} & \srci{32.3}{23.8,42.2} & $+9.4$ & \srci{59.4}{49.4,68.7} & \srci{32.3}{23.8,42.2} & $+27.1$ & \srci{55.2}{45.3,64.8} & \srci{22.9}{15.6,32.3} & $+32.3$ & $+22.9 \pm 12.0$ \\\addlinespace[1.5pt]
      clutter-pick-lift & \srci{22.9}{15.6,32.3} & \srci{13.5}{8.1,21.8} & $+9.4$ & \srci{19.8}{13.1,28.9} & \srci{14.6}{8.9,23.0} & $+5.2$ & \srci{13.5}{8.1,21.8} & \srci{11.5}{6.5,19.4} & $+2.1$ & $+5.6 \pm 3.7$ \\\addlinespace[1.5pt]
      multi-shelf-scan & \srci{17.7}{11.4,26.5} & \srci{0.0}{0.0,3.8} & $+17.7$ & \srci{17.7}{11.4,26.5} & \srci{3.1}{1.1,8.8} & $+14.6$ & \srci{9.4}{5.0,16.9} & \srci{7.3}{3.6,14.3} & $+2.1$ & $+11.5 \pm 8.3$ \\\addlinespace[1.5pt]
      \midrule
      \textit{suite mean} & \srci{47.1}{42.7,51.6} & \srci{38.8}{34.5,43.2} & $+8.3$ & \srci{54.6}{50.1,59.0} & \srci{32.7}{28.7,37.0} & $+21.9$ & \srci{46.7}{42.2,51.1} & \srci{23.8}{20.2,27.8} & $+22.9$ & $+17.7 \pm 8.1$ \\
      \bottomrule
    \end{tabular}}%
    }
\end{table}

\begin{table}[h!]
  \centering
  \caption{\textbf{GALT (s) of the multi-task $\pi_0$ headcam policies across the three training runs of Table~\ref{tab:seeds_sr}.} Median and inter-quartile range over successful episodes with a valid GALT detection; $n$: number of such episodes.}
  \label{tab:seeds_galt}
    {\scriptsize
    \setlength{\tabcolsep}{4pt}\renewcommand{\arraystretch}{1.1}
    \begin{tabular}{l@{\hskip 6pt} cc cc cc}
      \toprule
      & \multicolumn{2}{c}{run 1 (Table~\ref{tab:main_results})} & \multicolumn{2}{c}{run 2 (seed 123)} & \multicolumn{2}{c}{run 3 (seed 7)} \\
      \cmidrule(lr){2-3}\cmidrule(lr){4-5}\cmidrule(lr){6-7} & median [IQR] & $n$ & median [IQR] & $n$ & median [IQR] & $n$ \\
      \midrule
      conditional-pick & 2.70\,{\tiny[2.52,2.90]} & 79 & 2.40\,{\tiny[2.25,2.65]} & 76 & 2.70\,{\tiny[2.48,2.90]} & 63 \\
      wait-then-act & 2.65\,{\tiny[2.30,2.80]} & 61 & 2.60\,{\tiny[2.35,2.90]} & 85 & 2.60\,{\tiny[2.42,2.80]} & 79 \\
      clutter-pick-cube & 2.65\,{\tiny[2.12,2.83]} & 39 & 2.60\,{\tiny[2.20,2.90]} & 57 & 2.40\,{\tiny[2.27,2.75]} & 51 \\
      clutter-pick-lift & 2.55\,{\tiny[2.45,2.80]} & 21 & 2.50\,{\tiny[2.38,2.88]} & 19 & 2.65\,{\tiny[2.54,2.75]} & 10 \\
      multi-shelf-scan & 1.80\,{\tiny[1.54,2.17]} & 16 & 2.10\,{\tiny[1.83,2.31]} & 16 & 2.20\,{\tiny[1.76,2.61]} & 8 \\
      \midrule
      \textit{pooled} & 2.65\,{\tiny[2.34,2.85]} & 216 & 2.50\,{\tiny[2.30,2.80]} & 253 & 2.60\,{\tiny[2.40,2.80]} & 211 \\
      \bottomrule
    \end{tabular}}
\end{table}

\subsection{Wrist-Camera Ablation}
\label{app:wrist}

Both camera conditions receive the two wrist cameras in addition to the head or fixed camera (Section~\ref{sec:eval-paired}). To separate the contribution of active head control from wrist-based perception, we retrained the GR1T2 multi-task $\pi_0$ TAVIS-Head policies, and the single-task Diffusion Policy checkpoints for \textit{conditional-pick} and \textit{clutter-pick-cube}, without wrist cameras, i.e., with the head or the fixed camera as the only visual input. We evaluated them on the \textit{id} split with the same protocol as the main results (Table~\ref{tab:wrist}; one training run per condition).

Removing the wrist cameras lowers success in both conditions by a similar fraction: the multi-task suite mean drops from 47.1\% to 27.5\% for headcam and from 38.8\% to 22.5\% for fixedcam, leaving the headcam/fixedcam ratio unchanged (1.22$\times$ in both cases). The difference in percentage points shrinks ($+8.3$ to $+5.0$\,pp) as both conditions move toward the floor, and the largest per-task difference remains on \textit{conditional-pick} ($+24.0$\,pp without wrist cameras). For single-task Diffusion Policy, the gap on \textit{conditional-pick} widens from $+14.6$ to $+29.2$\,pp, because removing the wrist cameras lowers the fixedcam policy (33.3\% to 19.8\%) but not the headcam one (47.9\% to 49.0\%). Overall, wrist views carry information that both conditions use for manipulation, consistent with \citet{hsu2022vision}, while the headcam advantage remains when they are removed.

\begin{table}[h!]
  \centering
  \caption{\textbf{Wrist-camera ablation: success rates (\%) with and without wrist cameras.} GR1T2, \textit{id} split, 96 episodes per cell, with 95\% Wilson confidence intervals. With wrist cameras: the runs of the main results (Tables~\ref{tab:main_results} and~\ref{tab:main_results_singletask_diffpol}). Without wrist cameras: retrained with the head or fixed camera as the only visual input. $\Delta$: headcam minus fixedcam (pp).}
  \label{tab:wrist}

    \noindent\makebox[\textwidth][c]{%
    {\scriptsize
    \setlength{\tabcolsep}{3pt}\renewcommand{\arraystretch}{1.1}
    \begin{tabular}{l@{\hskip 6pt} ccc ccc}
      \toprule
      & \multicolumn{3}{c}{with wrist cameras} & \multicolumn{3}{c}{without wrist cameras} \\
      \cmidrule(lr){2-4}\cmidrule(lr){5-7} & head & fixed & $\Delta$ & head & fixed & $\Delta$ \\
      \midrule
      \multicolumn{7}{l}{\textbf{\color[HTML]{555555}multi-task $\pi_0$ (TAVIS-Head)}} \vspace{0.03in} \\
      conditional-pick & \srci{87.5}{79.4,92.7} & \srci{59.4}{49.4,68.7} & $+28.1$ & \srci{42.7}{33.3,52.7} & \srci{18.8}{12.2,27.7} & $+24.0$ \\\addlinespace[1.5pt]
      wait-then-act & \srci{65.6}{55.7,74.4} & \srci{88.5}{80.6,93.5} & $-22.9$ & \srci{60.4}{50.4,69.6} & \srci{64.6}{54.6,73.4} & $-4.2$ \\\addlinespace[1.5pt]
      clutter-pick-cube & \srci{41.7}{32.3,51.7} & \srci{32.3}{23.8,42.2} & $+9.4$ & \srci{19.8}{13.1,28.9} & \srci{17.7}{11.4,26.5} & $+2.1$ \\\addlinespace[1.5pt]
      clutter-pick-lift & \srci{22.9}{15.6,32.3} & \srci{13.5}{8.1,21.8} & $+9.4$ & \srci{8.3}{4.3,15.6} & \srci{11.5}{6.5,19.4} & $-3.1$ \\\addlinespace[1.5pt]
      multi-shelf-scan & \srci{17.7}{11.4,26.5} & \srci{0.0}{0.0,3.8} & $+17.7$ & \srci{6.2}{2.9,13.0} & \srci{0.0}{0.0,3.8} & $+6.2$ \\\addlinespace[1.5pt]
      \midrule
      \textit{suite mean} & \srci{47.1}{42.7,51.6} & \srci{38.8}{34.5,43.2} & $+8.3$ & \srci{27.5}{23.7,31.7} & \srci{22.5}{19.0,26.4} & $+5.0$ \\
      \midrule
      \multicolumn{7}{l}{\textbf{\color[HTML]{555555}single-task Diffusion Policy}} \vspace{0.03in} \\
      conditional-pick & \srci{47.9}{38.2,57.8} & \srci{33.3}{24.7,43.2} & $+14.6$ & \srci{49.0}{39.2,58.8} & \srci{19.8}{13.1,28.9} & $+29.2$ \\\addlinespace[1.5pt]
      clutter-pick-cube & \srci{42.7}{33.3,52.7} & \srci{30.2}{21.9,40.0} & $+12.5$ & \srci{15.6}{9.7,24.2} & \srci{6.2}{2.9,13.0} & $+9.4$ \\\addlinespace[1.5pt]
      \bottomrule
    \end{tabular}}%
    }
\end{table}

\clearpage

\section{Dataset Documentation}
\label{app:dataset_doc}

\noindent\textbf{Format and hosting.} The four TAVIS datasets (one per suite$\,\times\,$robot combination, totalling ${\sim}2200$ episodes and ${\sim}3$h of teleoperation, 800 episodes per robot in the head suite, 300 per robot in the hands one) are released as LeRobotDataset v3.0 repositories on the project Hugging Face organisation\footnote{\url{https://huggingface.co/tavis-benchmark}}, under a CC-BY-4.0 license. Each dataset includes synchronised head, fixed, left-wrist, and right-wrist RGB streams ($640\times 480$, MP4-encoded), full proprioceptive state, $19$-dimensional canonical actions, and per-episode language instructions where applicable. Episodes are further labelled with the Python class name (\texttt{task} field) for the corresponding task, so that single-task training is possible by filtering episodes from the published multi-task datasets.

\noindent\textbf{Maintenance.} The benchmark is maintained by the authors via the project code repository (\url{\repourl}) and the Hugging Face organisation. The authors commit to issue triage and PR review for at least two years post-publication. Future versions (additional tasks, robots, scene variations) will be released as new tagged versions in the repository.

\noindent\textbf{Reproducibility.} All training and evaluation scripts are publicly released alongside the datasets; pretrained $\pi_0$ multi-task checkpoints are hosted as separate model repositories on the same Hugging Face organisation.

\section{GALT (Gaze-Action Lead Time): Algorithm, Hyperparameters, and Validation}
\label{app:galt-algo}

\noindent\textbf{Algorithm overview.} We implement a single, sim-free GALT detector that consumes only the episode's commanded-action trajectory (Algorithm~\ref{alg:galt}). The action stream exposes neck joint targets, end-effector Cartesian targets, and gripper commands, which together are sufficient to identify the two events that define GALT: the latest gripper state change in the episode (anchor, $t^{\text{hand}}$) and the matching head fixation ($t^{\text{head}}$) found within a backward search window from the anchor. GALT is then $t^{\text{hand}} - t^{\text{head}}$, in seconds. A positive GALT indicates the head arrived at the task-relevant fixation before the gripper event, consistent with the active-vision pattern observed in teleoperation. The detector additionally validates that the gripper event was preceded by a real end-effector reach (i.e., a stable-to-motion transition before the anchor); episodes lacking such a reach are rejected as spurious. Because the detector relies on commanded actions rather than simulator state or external sensing, the metric transfers unchanged to real-robot deployments. For tasks with multiple grasp/release events per episode, the algorithm generalises trivially by iterating over all anchors and returning a list of per-event GALTs; the released implementation returns a single value anchored on the last gripper event, sufficient for the task suite reported here.

\noindent\textbf{Extensibility to other robots.} GALT is intentionally a \emph{family} of metrics rather than a single fixed algorithm, parameterised by which channels of the action vector represent arm end-effector positions, head joints, and gripper states. Our released implementation expresses this via a small \texttt{ActionLayout} specification mapping these channels to indices in the canonical $19$-dimensional action vector. Any robot whose policy outputs include or can compute arm-EE-position, head-joint, and gripper-scalar streams can plug into the same code. Users with their own datasets only need to construct the corresponding action trajectory to apply the detector as-is.

\noindent\textbf{Hyperparameter calibration.} All GALT hyperparameters (Table~\ref{tab:galt-hparams}) were calibrated once on teleoperation reference episodes and applied \emph{unchanged} to all policy rollouts reported in this paper. Differences between teleop and policy GALT distributions therefore reflect behavioural differences, not detector differences.

\noindent\textbf{Parameter rationale.} The two speed thresholds ($v_h, v_n$) distinguish ``moving'' from ``holding'' phases in the commanded trajectory; their values reflect typical teleoperation noise floors on EE pose (a few cm/s) and neck joints (a few deg/s). The persistence windows ($K_f$, $\tau_s$) suppress brief oscillations near the thresholds: $K_f$ requires a fixation to last at least $\sim\!80$\,ms before counting, and $\tau_s$ requires $\sim\!300$\,ms of prior stillness before a hand onset is declared, ensuring we pick up genuine stable-to-motion transitions rather than mid-reach micro-pauses. The search windows $(L, S)$ are generous enough to cover multi-second gaze-ahead patterns while cutting off at $3$\,s pre-anchor to avoid picking up fixations from earlier task phases. The refinement margin $r$ ($\sim\!3^\circ$) handles the smooth-deceleration case where the head reaches its fixation direction well before the velocity threshold is crossed; walking back in joint-position space recovers the true onset of fixation. Finally, $[\gamma_{\min}, \gamma_{\max}]$ discards pathological detections (e.g., gripper events unrelated to the task grasp) without changing the success-rate denominator: an outlier-flagged episode contributes to SR but not to the GALT distribution.

\noindent\textbf{Detection-rate validation.} On the teleoperation reference episodes (TAVIS-Head, $n\!=\!800$ per robot), the detector produces a valid GALT reading for $98.8\%$ of GR1T2 and $98.9\%$ of Reachy2 demonstrations at the native $60$\,Hz dataset rate (Table~\ref{tab:galt-validation}), confirming that the heuristic reliably captures the gaze-manipulation structure embedded in expert behaviour. At the $20$\,Hz policy-evaluation stride, detection rate drops to $85.6\%$ (GR1T2) and $93.0\%$ (Reachy2) -- entirely due to a tighter $\tau_s$ stability budget (only $6$ frames at $20$\,Hz vs $18$ at $60$\,Hz). Critically, the pooled median GALT itself is stable to within $\pm 20$\,ms across sampling rates, confirming that the metric is robust to evaluation stride and that the policy-rollout GALT distributions reported in the main text are not biased by the rate change.

\begin{table}[h]
\centering
\caption{\textbf{GALT detection-rate validation on teleoperation reference episodes.} Pooled across the $5$ TAVIS-Head tasks, $n\!=\!800$ episodes per robot. Mean and median GALT shown only for valid (non-rejected) episodes.}
\label{tab:galt-validation}
\small
\renewcommand{\arraystretch}{1.1}
\begin{tabular}{@{}lccc@{}}
\toprule
& Detection rate & Pooled mean GALT (s) & Pooled median GALT (s) \\
\midrule
GR1T2 @ $60$\,Hz (native) & $790 / 800$ ($98.8\%$) & $2.57$ & $2.57$ \\
GR1T2 @ $20$\,Hz (eval stride) & $685 / 800$ ($85.6\%$) & $2.55$ & $2.55$ \\
Reachy2 @ $60$\,Hz (native) & $791 / 800$ ($98.9\%$) & $2.10$ & $2.10$ \\
Reachy2 @ $20$\,Hz (eval stride) & $744 / 800$ ($93.0\%$) & $2.11$ & $2.10$ \\
\bottomrule
\end{tabular}
\end{table}

\noindent\textbf{Sensitivity to detector parameters.} To test whether the reported lead times depend on the chosen thresholds, we varied each of the six main detector parameters one at a time over roughly $\pm2\times$ its default, keeping the others at their defaults (29 configurations per episode set). We applied the sweep to teleoperation demonstrations (60\,Hz) and to successful policy rollouts of the GR1T2 headcam checkpoints (20\,Hz; re-run from the same checkpoints with action logging), as summarised in Table~\ref{tab:galt-sweep}. Except for the arrival margin $r$, no parameter moves the median lead by more than 0.06\,s on the demonstrations or 0.25\,s on the rollouts. The arrival margin moves it more (1.48--3.00\,s across episode sets), by construction: a looser margin declares head arrival earlier and therefore reports a longer lead. In every configuration, at least 75\% of detected episodes keep a lead of 1\,s or more. Detection rates vary more across settings at the 20\,Hz rollout rate than on the 60\,Hz demonstrations; when a configuration rejects an episode, the detector returns no value rather than a different one. \textit{multi-shelf-scan} is omitted, as only 11 successful rollouts were available.

\begin{table}[h!]
\centering
\caption{\textbf{GALT sensitivity to detector parameters.} Median GALT (s) when each parameter is varied alone over the listed range, with all others at their defaults (Table~\ref{tab:galt-hparams}); a single value means the median did not change. Demonstrations: teleoperation episodes at 60\,Hz ($n=100$ per task). Rollouts: successful episodes of the GR1T2 headcam checkpoints at 20\,Hz (multi-task $\pi_0$: $n=74$, 35 and 14; single-task Diffusion Policy: $n=46$). Last row: minimum, over the 29 configurations, of the share of detected episodes with GALT $\geq 1$\,s.}
\label{tab:galt-sweep}
{\scriptsize
\setlength{\tabcolsep}{3pt}\renewcommand{\arraystretch}{1.15}
\begin{tabular}{@{}lll cc cccc@{}}
\toprule
 & & & \multicolumn{2}{c}{demonstrations (60\,Hz)} & \multicolumn{4}{c}{policy rollouts (20\,Hz)} \\
\cmidrule(lr){4-5}\cmidrule(lr){6-9}
parameter & default & range & \makecell{conditional-\\pick} & \makecell{clutter-\\pick-cube} & \makecell{$\pi_0$\\conditional-pick} & \makecell{$\pi_0$\\clutter-pick-cube} & \makecell{$\pi_0$\\clutter-pick-lift} & \makecell{DP\\conditional-pick} \\
\midrule
      $v_h$ (m/s) & 0.05 & 0.02--0.1 & 2.52--2.58 & 2.62--2.68 & 2.62--2.70 & 2.27--2.40 & 2.85 & 2.38--2.50 \\
      $v_n$ (rad/s) & 0.1 & 0.05--0.2 & 2.55--2.58 & 2.67--2.68 & 2.70--2.77 & 2.40--2.60 & 2.83--2.95 & 2.30--2.40 \\
      $K_f$ (s) & 0.08 & 0.04--0.16 & 2.58 & 2.67--2.68 & 2.70 & 2.35--2.55 & 2.80--2.85 & 2.35--2.38 \\
      $\tau_s$ (s) & 0.3 & 0.15--0.6 & 2.58 & 2.68 & 2.70 & 2.40--2.55 & 2.70--2.85 & 2.35--2.60 \\
      $r$ (rad) & 0.05 & 0.025--0.1 & 2.27--2.68 & 2.36--2.80 & 1.90--2.88 & 1.55--2.65 & 1.48--3.00 & 1.50--2.80 \\
      $L$ (s) & 3 & 2--5 & 2.58 & 2.68 & 2.70 & 2.40 & 2.85 & 2.38 \\
      \midrule
      \multicolumn{3}{l}{min.\ \% of episodes with GALT $\geq 1$\,s} & 91 & 94 & 93 & 83 & 79 & 75 \\
\bottomrule
\end{tabular}}
\end{table}

\FloatBarrier
\noindent\textbf{Validation against gaze-target geometry.} GALT infers head arrival from commanded neck motion alone. To check that the detected fixation is directed at the task-relevant object, we re-ran the GR1T2 multi-task $\pi_0$ headcam policy on the five TAVIS-Head tasks, and the single-task Diffusion Policy on \textit{conditional-pick}, logging the head-camera pose and the position of every object at each control step. The camera pose is derived from the head link to which the camera is rigidly mounted (optical axis along the link's forward axis, 0.1\,m mount offset). At the head-arrival time returned by the unchanged GALT detector, we measure (i) the angle between the optical axis and the direction to the target; (ii) the same angle at episode start, before any head motion, as a reference for the scene geometry; and (iii) whether the target is the object nearest to the optical axis among all $N$ task objects, for which chance is $1/N$. Only geometry is taken from these re-runs; success rates are those of the main results.

Pooled over the five tasks ($n=188$ successful $\pi_0$ episodes), the optical axis lies a median of 7.5$^\circ$ [IQR 4.3, 9.6] from the target at the detected head arrival, against 52.0$^\circ$ at episode start, and within the central half of the camera's field of view ($17.5^\circ$) in 99\% of episodes (Table~\ref{tab:galt-geometry}). Pooled over the same episodes, the target is the object nearest to the optical axis in 81\% of them, against 34\% expected by chance; per task, this share ranges from 57\% to 100\%, against chance levels of 20--50\% that depend on the number of objects. Diffusion Policy on \textit{conditional-pick} gives the same picture. The residual offset is consistent with head-directed gaze, whose precision is bounded by neck control rather than by foveal targeting: on \textit{conditional-pick}, the 4.9$^\circ$ median offset is about 1.4 times the target's angular radius (3.4$^\circ$), a lateral miss of 4.2\,cm at 0.49\,m range. The lowest nearest-object rate is on \textit{wait-then-act} (78\% against 50\% chance), where gaze visits the status light before the object: in the remaining episodes, the detected fixation is closer to the light, as the detector's backward search can settle on the cue fixation in this cue-then-target task.

\begin{table}[h!]
\centering
\caption{\textbf{Gaze-target geometry at the GALT-detected head arrival.} GR1T2 headcam policies, \textit{id} split, successful episodes with a valid GALT detection. Angles between the head camera's optical axis and the direction to the target, in degrees (median [IQR]); central field: within $17.5^\circ$, half of the field-of-view half-angle; nearest object: share of episodes in which the target is the object nearest to the optical axis, with chance level $1/N$ for $N$ task objects.}
\label{tab:galt-geometry}
{\small
\setlength{\tabcolsep}{4pt}\renewcommand{\arraystretch}{1.1}
\begin{tabular}{@{}llr ccccc@{}}
\toprule
 & & & \multicolumn{2}{c}{angle to target ($^\circ$)} & & \multicolumn{2}{c}{nearest object (\%)} \\
\cmidrule(lr){4-5}\cmidrule(lr){7-8}
task & policy & $n$ & at head arrival & at start & central field (\%) & observed & chance \\
\midrule
      conditional-pick & $\pi_0$ & 70 & 4.9 [3.8, 8.8] & 52.8 & 100 & 93 & 33 \\
      wait-then-act & $\pi_0$ & 58 & 7.9 [4.4, 9.5] & 50.7 & 100 & 78 & 50 \\
      clutter-pick-cube & $\pi_0$ & 35 & 8.8 [6.2, 10.2] & 55.5 & 100 & 57 & 20 \\
      clutter-pick-lift & $\pi_0$ & 14 & 10.0 [8.2, 12.4] & 52.8 & 100 & 86 & 20 \\
      multi-shelf-scan & $\pi_0$ & 11 & 7.7 [4.0, 8.8] & 25.7 & 91 & 91 & 20 \\
      \midrule
      conditional-pick & DP & 44 & 4.8 [3.1, 6.5] & 55.0 & 100 & 100 & 33 \\
\bottomrule
\end{tabular}}
\end{table}

\begin{algorithm}[t]
\DontPrintSemicolon
\SetAlgoLined
\small
\SetKwInOut{Input}{Input}\SetKwInOut{Output}{Output}

\Input{Action trajectory $A \in \mathbb{R}^{T \times 19}$, sampling rate $f$\,Hz, hyperparameters (Table~\ref{tab:galt-hparams}).}
\Output{GALT $g \in \mathbb{R}$ in seconds, or stop with reason code.}

$\dot q_n \gets$ per-step neck angular speed\;
\ForEach{arm $a \in \{L, R\}$}{
$\dot p_a \gets$ per-step end-effector linear speed; \quad
$E_a \gets$ gripper-command sign changes\;
$t^{\text{hand}}_a \gets \max E_a$ \tcp*{gripper anchor}
$t^{\text{onset}}_a \gets$ end of latest $\geq \tau_s f$-step run with $\dot p_a < v_h$, before $t^{\text{hand}}_a$\;
\lIf{none}{\textbf{skip} (\texttt{no\_hand\_onset})}
$t^{\text{head}}_a \gets$ start of nearest $\geq K_f f$-step run with $\dot q_n < v_n$ within $[t^{\text{hand}}_a-Lf,\, t^{\text{hand}}_a+Sf]$, refined back in joint-position space within margin $r$\;
\lIf{none}{\textbf{skip} (\texttt{no\_fixation})}
$g_a \gets (t^{\text{hand}}_a - t^{\text{head}}_a) / f$\;
\lIf{$g_a \notin [\gamma_{\min}, \gamma_{\max}]$}{\textbf{skip} (\texttt{outlier})}
arm $a$ valid with value $g_a$\;
}
\lIf{both arms valid}{\Return \texttt{ambiguous\_arms}}
\lIf{no arm valid}{\Return most informative skip reason}
\Return GALT $= g_{a^*}$ for the unique valid arm $a^*$\;
\caption{\textbf{GALT (Gaze-Action Lead Time) detector.} Operates on the canonical 19-D action trajectory only; anchor is the last gripper-command sign change, and a valid result requires unique-arm validation.}
\label{alg:galt}
\end{algorithm}

\begin{table}[h!]
\centering
\small
\caption{\textbf{GALT detector hyperparameters.} Calibrated once on the teleoperation reference episodes and applied unchanged to all policy rollouts. Code-side variable names: $v_h$=\texttt{v\_hand\_thresh}, $v_n$=\texttt{v\_sac\_thresh}, $K_f$=\texttt{K\_fix\_s}, $\tau_s$=\texttt{min\_stable\_for\_onset\_s}, $L$=\texttt{lookback\_s}, $S$=\texttt{forward\_slack\_s}, $r$=\texttt{arrival\_margin\_rad}, $\gamma_{\min,\max}$=\texttt{outlier\_min/max\_s}.}
\label{tab:galt-hparams}
\begin{tabular}{@{}lllp{0.50\linewidth}@{}}
\toprule
Symbol & Value & Unit & Description \\
\midrule
\multicolumn{4}{@{}l}{\textit{Speed thresholds} (moving phase vs~stationary phase)} \\
$v_h$ & $0.05$ & m/s & End-effector commanded linear-speed floor. \\
$v_n$ & $0.10$ & rad/s & Neck commanded angular-speed floor. \\
\addlinespace
\multicolumn{4}{@{}l}{\textit{Persistence / stability windows}} \\
$K_f$ & $0.080$ & s & Minimum duration below $v_n$ to qualify as a fixation. \\
$\tau_s$ & $0.300$ & s & Minimum stable (below-$v_h$) duration preceding a hand onset. \\
\addlinespace
\multicolumn{4}{@{}l}{\textit{Search windows around the gripper anchor}} \\
$L$ & $3.0$ & s & Backward lookback horizon for head-arrival candidates. \\
$S$ & $0.5$ & s & Post-anchor slack for head-arrival candidates. \\
\addlinespace
\multicolumn{4}{@{}l}{\textit{Arrival refinement}} \\
$r$ & $0.05$ & rad ($\approx\!2.9^\circ$) & $L_\infty$ neck-joint margin defining ``at final fixation''. \\
\addlinespace
\multicolumn{4}{@{}l}{\textit{Outlier bounds on the final GALT value}} \\
$\gamma_{\min}$ & $-0.5$ & s & Below this $\Rightarrow$ \texttt{outlier\_low}. \\
$\gamma_{\max}$ & $~4.0$ & s & Above this $\Rightarrow$ \texttt{outlier\_high}. \\
\bottomrule
\end{tabular}
\end{table}

\clearpage
\section{Gaze Behaviour Statistics}
\label{app:gaze-stats}

Beyond the timing captured by GALT, we characterise how policies move the head, using the commanded neck trajectories of the logged policy rollouts of Appendix~\ref{app:galt-algo} (successful episodes, 20\,Hz) and of the teleoperation demonstrations. A head movement is a rising edge of the commanded neck angular speed above $v_n$, the threshold used by the GALT detector (Table~\ref{tab:galt-hparams}); head travel is the summed absolute change of the neck joint angles over an episode. Because the number of detected movements depends on the sampling rate, we compare the policies with demonstrations downsampled to the same 20\,Hz rate, and report native-rate values for reference.

\begin{figure}[h!]
  \centering
  \includegraphics[width=\textwidth]{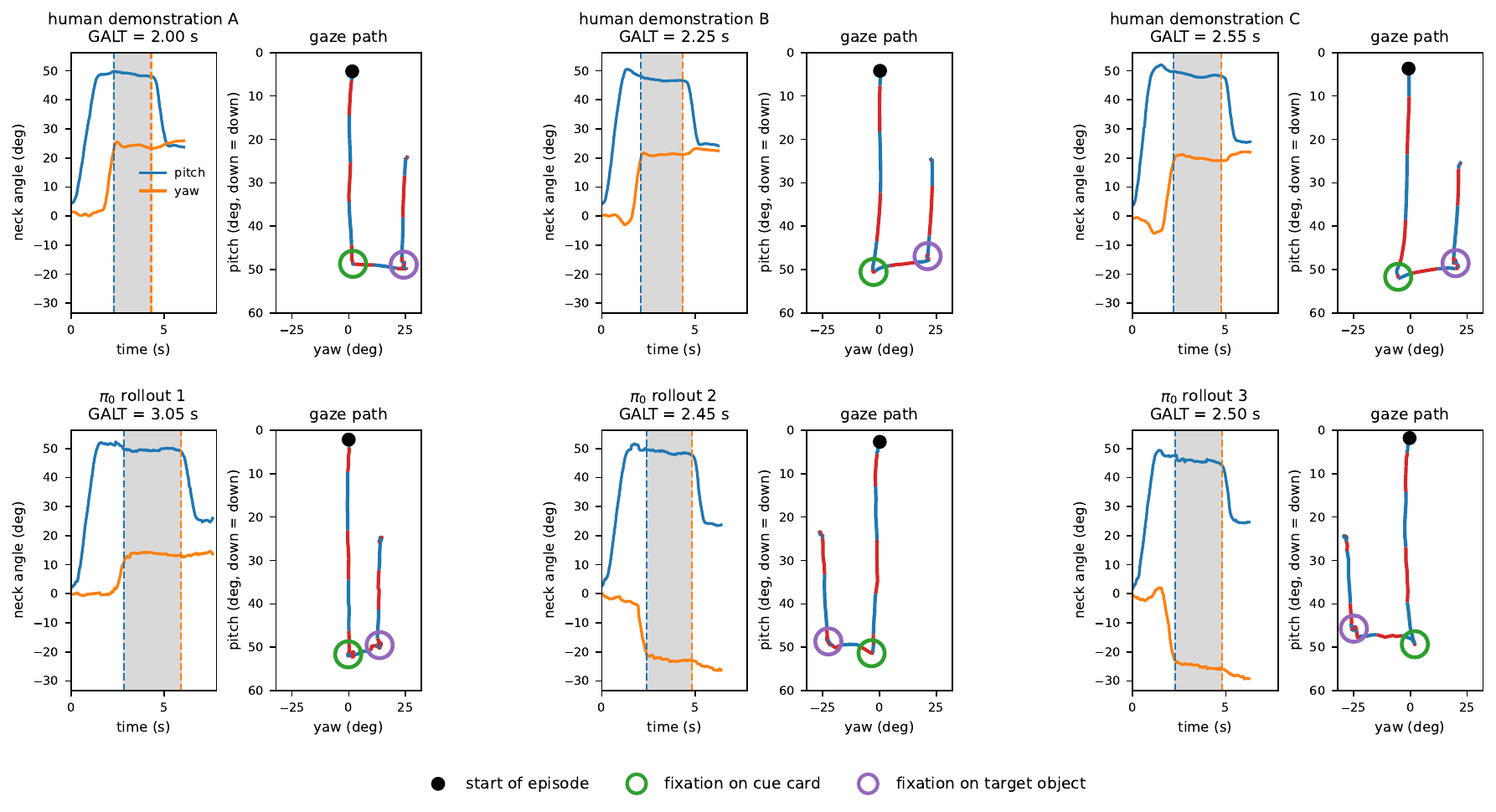}
  \caption{\textbf{Head trajectories on \textit{conditional-pick} (GR1T2): human demonstrations (top) and multi-task $\pi_0$ rollouts (bottom).} Three randomly selected episodes per source; columns are not paired. In each episode, the head first pitches down to the colored cue card, then turns left or right toward the target object that the card indicates, holds that fixation until the grasp, and pitches back up while the object is lifted. For each episode, left: commanded neck pitch and yaw over time, with the GALT interval shaded from head arrival (blue dashed line) to the grasp (orange dashed line). Right: the gaze path in yaw--pitch space (pitch axis inverted, so that looking down points down), starting at the black dot and drawn in alternating 0.25\,s segments so that segment length reflects angular speed. Circles mark the fixation on the cue card (green; the last stable head pose before turning toward the target) and on the target object (purple; mean head pose over the GALT interval). Demonstrations are downsampled to the policy rate of 20\,Hz.}
  \label{fig:gaze-traj}
\end{figure}

Table~\ref{tab:gaze-moves} decomposes head travel into the number of movements and the travel per movement. On the two tasks with matched demonstrations, the demonstrations contain about 3 movements of 34--37$^\circ$ each, whereas $\pi_0$ makes 12--14 movements of about 12$^\circ$. The extra movements are small, and total travel is only about 1.5$\times$ that of the demonstrations. Figure~\ref{fig:gaze-traj} shows example trajectories on \textit{conditional-pick}: policies follow the same cue-then-target gaze path as the demonstrations.

To locate the extra movements in time, we compute the phase of each movement onset within the 8-step action chunk that the policies execute per inference call. Onsets are far from uniform over the 8 phases (12.5\% each under uniformity): a single phase holds 37--72\% of them, in both $\pi_0$ and Diffusion Policy (last column of Table~\ref{tab:gaze-moves}). Head motion is therefore smooth within a chunk and discontinuous at chunk boundaries; if the extra movements were threshold noise, their onsets would be spread uniformly across phases. The number of detected movements depends on the speed threshold, whereas total head travel does not.

\begin{table}[h!]
\centering
\caption{\textbf{Head movements and head travel per episode.} GR1T2, \textit{id} split, successful episodes (medians). A movement is a rising edge of the commanded neck angular speed above $v_n$; travel is in degrees. Last column: share of movement onsets falling on the most frequent of the 8 phases of the executed action chunk (12.5\% if uniform).}
\label{tab:gaze-moves}
{\small
\setlength{\tabcolsep}{4pt}\renewcommand{\arraystretch}{1.1}
\begin{tabular}{@{}llcr cccc@{}}
\toprule
source & task & rate & $n$ & movements & \makecell{travel /\\movement} & travel & \makecell{onsets at peak\\chunk phase (\%)} \\
\midrule
      demonstrations & conditional-pick & 20 Hz & 100 & 3 & 34.1 & 108 & -- \\
      demonstrations & clutter-pick-cube & 20 Hz & 100 & 3 & 37.1 & 91 & -- \\
      \midrule
      $\pi_0$ & conditional-pick & 20 Hz & 74 & 14 & 11.9 & 167 & 37 \\
      $\pi_0$ & clutter-pick-cube & 20 Hz & 35 & 12 & 12.0 & 140 & 42 \\
      $\pi_0$ & clutter-pick-lift & 20 Hz & 14 & 11 & 11.4 & 129 & 42 \\
      $\pi_0$ & wait-then-act & 20 Hz & 61 & 16 & 9.4 & 155 & 38 \\
      $\pi_0$ & multi-shelf-scan & 20 Hz & 11 & 8 & 7.1 & 53 & 49 \\
      Diffusion Policy & conditional-pick & 20 Hz & 46 & 10 & 13.4 & 141 & 72 \\
      \midrule
      \multicolumn{8}{l}{\textit{demonstrations at their native 60\,Hz rate, for reference}} \\
      demonstrations & conditional-pick & 60 Hz & 100 & 8 & 14.7 & 110 & -- \\
      demonstrations & clutter-pick-cube & 60 Hz & 100 & 6 & 14.2 & 92 & -- \\
\bottomrule
\end{tabular}}
\end{table}

\end{document}